\documentclass[preprint,review,12pt]{elsarticle}

\usepackage{amssymb}
\usepackage{amsmath}
\usepackage{graphicx}
\usepackage{subfigure}
\usepackage{float}
\usepackage{threeparttable}
\usepackage{verbatim}

\usepackage{setspace} 
\usepackage{amsfonts}
\usepackage{algorithm}
\usepackage{algorithmic}
\usepackage{indentfirst}
\usepackage{float}

\usepackage{url}
\usepackage{caption}
\usepackage{bbding} 

\usepackage{colortbl} 
\usepackage{xcolor}
\usepackage{array}
\usepackage{makecell}
\usepackage{multirow}
\usepackage{color}

\newtheorem{theorem}{Theorem}

\newtheorem{remark}[theorem]{Remark}  

\usepackage{geometry}
 \usepackage{array}
 \usepackage{booktabs}
 \usepackage{graphicx}
 \usepackage{subfigure}
 \usepackage{float}
 \usepackage{threeparttable}
 \usepackage{verbatim}
 \usepackage{amsfonts}
 \usepackage{algorithm}
 \usepackage{algorithmic}
 \usepackage{indentfirst}
 \usepackage{float}
 
 \usepackage{url}
 \usepackage{caption}
 \usepackage{bbding} 

 \usepackage{colortbl} 
 \usepackage{xcolor}
 \usepackage{array}
 \usepackage{color}
 
 \usepackage{bbm}
 \usepackage[english]{babel}
\usepackage{xspace}
\usepackage{csquotes}
\usepackage{makecell}
\usepackage{multirow}

\usepackage{pifont}

\newtheorem{definition}{Definition}

\begin{document}

\begin{frontmatter}

\title{Towards Accurate and Interpretable Time-series Forecasting: A Polynomial Learning Approach}

\author[author1,author2]{Bo Liu}  
\author[author1]{Shao-Bo Lin}
\author[author3]{Changmiao Wang}  
\author[author1]{Xiaotong Liu \corref{cor1}}

\address[author1]{Center for Intelligent Decision-Making and Machine Learning, School of Management, Xi'an Jiaotong University, Xi'an, China}
\address[author2]{The 39th Research Institute of China Electronics Technology Group Corporation, Xi'an, China}
\address[author3]{Shenzhen Research Institute of Big Data, ShenZhen, China}

\cortext[cor1]{Corresponding author: ariesoomoon@gmail.com}

\begin{abstract}

Time series forecasting enables early warning and has driven asset performance management from traditional planned maintenance to predictive maintenance. However, the lack of interpretability in forecasting methods  undermines users’ trust and complicates debugging for developers. Consequently, interpretable time-series forecasting has attracted increasing research attention.
Nevertheless, existing methods suffer from several limitations, including insufficient modeling of temporal dependencies, lack of feature-level interpretability to support early warning, and difficulty in simultaneously achieving the accuracy and interpretability.
This paper proposes the interpretable polynomial learning (IPL) method, which integrates interpretability into the model structure by explicitly modeling original features and their interactions of arbitrary order through polynomial representations. This design preserves temporal dependencies, provides feature-level interpretability, and offers a flexible trade-off between prediction accuracy and interpretability by adjusting the polynomial degree.
We evaluate IPL on simulated and Bitcoin price data, showing that it achieves high prediction accuracy with superior interpretability compared with widely used explainability methods. Experiments on field-collected antenna data further demonstrate that IPL yields simpler and more efficient early warning mechanisms.

\end{abstract}

\begin{keyword}
time-series forecasting; interpretability; early warning; polynomial learning

\end{keyword}

\end{frontmatter}


\section{Introduction}\label{introduction}

Time-series data sequentially record the evolution of system states, encompassing stock price fluctuations in financial markets, sensor readings from industrial equipment, and ECG monitoring in healthcare
\citep{zhang2024generalized, taieb2021hierarchical, bertani2025joint}.  
Leveraging the inherent temporal dependencies in time-series data, future states can be inferred through the identification of recurrent historical patterns; 
consequently, time-series forecasting is widely applied across critical domains such as healthcare early warning systems, engineering predictive maintenance, and financial market monitoring \citep{mori2019early, guo2025explainable}.
Indeed, in asset performance management (APM), the focus has shifted from traditional planned maintenance to time-series-forecasting-driven  predictive maintenance\footnote{https://www.gelonghui.com/p/916603},
 with the value of an accurate early warning of a major equipment failure exceeding $\$100,000$ in many industries\footnote{https://iot-analytics.com/product/predictive-maintenance-asset-performance-market-report-2023-2028/}.
In 2024, the global predictive maintenance market was valued at $\$$10.93 billion and is projected to grow to $\$$70.73 billion by 2032, with a compound annual growth rate of 26.5\% during the forecast period\footnote{https://www.fortunebusinessinsights.com/press-release/predictive-maintenance-market-9602}. 

Existing time-series–based early warning systems (e.g., OpenText, Falkonry, DolphinDB, and Artesis) predominantly rely on two paradigms: (1) rule-based alerting driven by human experience, and (2) anomaly scoring that triggers alarms based on the deviation of key indicators from their normal states. Neither paradigm is truly based on data-driven predictive models, and both struggle to ensure that early warnings actually correspond to true anomalies. Consequently, according to IoT Analytics, the accuracy of current early warning systems largely remains below 50\%, which further leads to alarm fatigue and undermines operators’ trust.
Meanwhile, existing time-series forecasting methods—particularly deep learning–based approaches such as RNNs, CNNs, and Transformers \citep{hochreiter1997long, vaswani2017attention, bai2018empirical, salinas2020deepar, lim2021time}—have achieved state-of-the-art prediction performance; however, their black-box nature makes it difficult to interpret how predictions are generated.
Beyond compelling prediction performance, it is ultimately interpretability—particularly feature-level interpretability, which clarifies the contribution of individual features to predictions—that enables these models to support trustworthy, actionable, and targeted interventions  following warnings \citep{kumar2025deepsecure, senoner2022using, bouazizi2024enhancing, guo2025explainable}.
For instance, in industrial settings, effective early warning requires not only anomaly alerts but also precise identification of the sensor readings indicating potential faults, enabling targeted repairs instead of extensive disassembly. 

Lacking such feature-level interpretability 
undermines user trust in time-series model outputs and complicates debugging for developers, thereby significantly hindering their real-world deployment in early-warning applications and raising critical legal, ethical, and equity concerns \citep{lim2021temporal, kumar2025deepsecure, wang2020deep, hen2025spatio, bauer2023expl, bauer2024mirror}.
A striking example of poor interpretability in the financial sector is the 2010 ``Flash Crash,'' where the U.S. stock market plunged nearly 1,000 points in minutes, triggered by trading systems executing massive sell-offs based on time-series patterns that could not be interpreted in real time. The event erased approximately 1 trillion in market value, ultimately undermining confidence in algorithmic trading systems.

Given the importance of interpretability for time-series–based early warning systems, significant efforts have been devoted to developing interpretable time-series forecasting methods.
These include approaches integrated with deep learning architectures, such as backpropagation-based methods \citep{oviedo2019fast}
and perturbation-based methods \citep{kashiparekh2019convtimenet} for convolutional neural networks, as well as attention mechanisms for recurrent neural networks \citep{choi2016retain}, and models specifically designed for time-series forecasting, such as the classical ARIMAX model \citep{box2015time}, along with more recent Symbolic Aggregate approXimation (SAX)–based methods \citep{nguyen2018interpretable}, fuzzy logic approaches \citep{wang2020deepfuzzy}, and shapelet-based models \citep{li2021shapenet}. 
 Further model-agnostic interpretability techniques, including extensions of LIME and SHAP for time-series data, also contribute to this growing field \citep{zhao2023interpretation}. 
 Comprehensive surveys of interpretable time-series forecasting methods are available for further reference \citep{lim2021time, zhao2023interpretation}.

Despite these advances, a notable gap remains between existing time-series interpretability methods and the practical demands of time-series–based early-warning applications. This gap manifests in three key aspects:
(1) \textit{Neglect of temporal dependencies in post‑hoc explanations}: Time-series data exhibit strong temporal dependencies \citep{
sun2022distributed}, and early-warning applications require explanations that clarify how specific temporal features influence final predictions. However, existing post‑hoc methods often fail to meet this requirement, yielding unreliable explanations: LIME‑based methods independently construct surrogate models for each data point \citep{ribeiro2016should}, and SHAP‑based methods evaluate features independently across neighboring time steps \citep{lundberg2017unified}.
(2) \textit{Limited feature‑level interpretability in deep learning methods}: 
Interpretability approaches  for deep learning models—such as those highlighting influential sub-sequences, attention weights, or salient temporal regions—primarily explain when critical information occurs in the input sequence, but provide limited insight into which features drive the prediction and thus lack the feature-level interpretability required for early-warning applications \citep{oviedo2019fast, kashiparekh2019convtimenet, choi2016retain}.
Moreover, these methods typically ignore the impact of feature interactions on the prediction outcomes \citep{guo2025explainable}.
(3) \textit{Accuracy--interpretability trade-off in inherently interpretable methods}: 
Inherently interpretable models such as ARIMAX provide feature-level interpretability but generally underperform deep learning--based approaches in prediction accuracy \citep{lim2021time, vaswani2017attention}, making it difficult to meet the dual requirements of accuracy and interpretability in early-warning applications.


To bridge this gap, we propose an interpretable polynomial learning (IPL) method for time-series forecasting. This method incorporates explicit polynomial structures, enabling direct interpretation of both individual features and their interaction effects on predictions. We compare IPL with widely used interpretability methods, including ARIMAX, LIME, and SHAP, and demonstrate that IPL outperforms these baselines not only in identifying feature contributions, but also in prediction accuracy and computational efficiency—attributes that are critical for real-time early-warning applications. Through perturbation analysis, we show that controlled modifications to input features lead to corresponding changes in prediction outcomes, thereby reflecting the importance of the perturbed features and validating IPL’s ability to quantify feature contributions to predictions. Furthermore, we evaluate IPL on two real-world datasets: Bitcoin market data and field-collected antenna data. For the latter, we construct an interpretability-driven early-warning mechanism, and experimental results demonstrate that the IPL-based warning mechanism achieves superior efficiency and prediction accuracy.



\begin{figure}[H]
	\centering
	\vspace{0.1in}
	\includegraphics[scale=0.35]{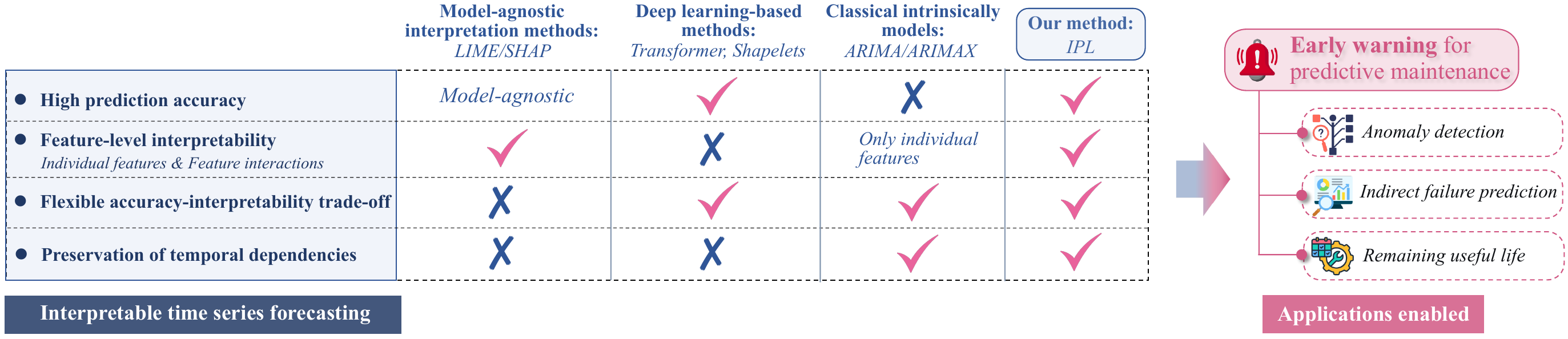}
        	\caption{A comparison of different interpretable time-series forecasting
methods}
	\label{Fig:compare_method}  
	\vspace{-0.3cm}
\end{figure}

This paper makes two main contributions. First, we contribute to the literature on interpretable time-series forecasting by proposing the IPL approach. By modeling input features through polynomial structures, IPL preserves temporal dependencies and provides feature-level interpretability.
Through a tunable polynomial order, IPL enables flexible adjustment of the accuracy–interpretability trade-off, allowing the selection of an order that simultaneously achieves high accuracy and feature-level interpretability, concretely implementing the flexible explainability policy suggested by \citep{mohammadi2025regulating}.  As illustrated in Figure~\ref{Fig:compare_method}, IPL uniquely integrates high accuracy, interpretability suitable for early warning, and a tunable accuracy–interpretability trade-off. Second, we contribute to the field of early warning in time-series analysis \citep{mori2019early} by demonstrating that, beyond individual features, feature interactions are also critical for prediction outcomes. Numerical results confirm that IPL effectively identifies the contributions of features—both individual and interactive—to predictions, outperforming inherently interpretable methods such as ARIMAX, which focus solely on individual features.

The remainder of this paper is organized as follows. Section 2 reviews related work on interpretable time-series forecasting. Section 3 introduces the proposed interpretable polynomial learning method. Sections 4, 5, and 6 present numerical experiments conducted on simulated data, real-world Bitcoin price data, and field-collected antenna data to validate the proposed method. Section 7 concludes the paper.

\section{Related work}

The growing importance of interpretable time-series forecasting is driven by the need to build trust, ensure accountability, and facilitate model debugging in high-stakes applications \citep{mueller2020longitudinal, lim2021temporal, kumar2025deepsecure, bertani2025joint, shao2025revisiting}. Rather than adhering to the conventional dichotomy of post-hoc versus ante-hoc interpretability, this review follows the developmental trajectory of time-series interpretability methods, aiming to clarify the gap between existing approaches and the practical requirements of early-warning applications.

The first category comprises classical intrinsically interpretable methods specifically designed for time series \citep{victor2016autoregressive, stock2001vector, box2015time}.
A representative example is ARIMAX, which achieves feature-level interpretability by directly incorporating original input features into the forecasting model. However, these approaches focus solely on individual input features and typically ignore interactions among them. Moreover, being predominantly linear or only mildly nonlinear, they often struggle to capture the complex patterns that modern deep learning architectures can learn \citep{lim2021time, vaswani2017attention}, leading to a trade-off where interpretability comes at the cost of prediction accuracy.

The second category includes model-agnostic post-hoc interpretability techniques, such as LIME-based \citep{ribeiro2016should, sivill2022limesegment} and SHAP-based \citep{lundberg2017unified, eun2025vector} methods. These techniques are typically applied to pre-trained black-box models and rely on either approximating the model behavior with a surrogate interpretable model (e.g., LIME) or decomposing predictions into feature-level attributions (e.g., SHAP).  While effective in tabular data settings \citep{senoner2022using} and capable of leveraging the high accuracy of complex pre-trained models, their application to time series faces significant limitations.  Specifically, LIME-based perturbations may disrupt the intrinsic dynamical structure of time series, potentially generating unrealistic synthetic samples and leading to unreliable explanations \citep{crabbe2021explaining}. 
Similarly, SHAP treats features at neighboring time steps as independent, which contradicts the inherent temporal autocorrelation of time series and undermines explanation fidelity. In addition, the computational cost of Shapley value estimation grows exponentially with the number of features, rendering SHAP impractical for long time series \citep{lundberg2017unified}.

The third category encompasses deep learning-based interpretability methods. These methods typically obtain explanations derived from sub-sequences, attention weights, or salient temporal regions \citep{oviedo2019fast, kashiparekh2019convtimenet, choi2016retain}, emphasizing temporal importance—when certain information is important or where the model focuses within the input sequence—rather than providing feature-level interpretability of the original input features or explaining how their interactions drive the prediction. In practical early-warning applications, only explanations grounded in observable and controllable input features can offer decision-makers intuitive and actionable insights.

In summary, existing approaches to interpretable time-series forecasting suffer from three major limitations: (1) classical linear models are constrained in prediction accuracy; (2) model-agnostic methods disrupt temporal structures and yield explanations with limited reliability; and (3) deep learning-based interpretability methods primarily focus on derived temporal representations rather than the original input features. Furthermore, these approaches generally overlook feature interactions and lack the flexibility to balance prediction accuracy and interpretability.

To address these challenges, we propose the interpretable polynomial learning (IPL) method. IPL tightly couples interpretability with model structure by explicitly modeling the original input features and their interactions of arbitrary order through polynomial representations, thereby preserving temporal dependencies while providing feature-level interpretability for both individual features and their interactions.
By adjusting the polynomial degree, IPL offers a flexible trade-off between prediction accuracy and interpretability. Experimental results demonstrate that IPL achieves high  accuracy while maintaining low computational cost, making it well suited for real-time early-warning applications.

\section{Interpretable polynomial learning for time series}

In this section, we first introduce feature-level interpretability, which serves as the motivation for our proposed method. We then present the interpretability imperatives in time-series forecasting, followed by a detailed description of the interpretable polynomial learning method and its specific algorithmic implementation for time series.

\subsection{Feature interpretability: dual focus on individual features and feature interactions}

Existing research on time-series interpretability has primarily focused on individual feature-based explanations while largely overlooking feature interactions \citep{guo2025explainable}. However, recent studies suggest that such explanations often fail to capture the intricate relationships that AI models exploit to generate accurate predictions.
In healthcare, for instance, drug interactions can lead to unexpected synergistic or antagonistic effects that are closely tied to a patient's condition \citep{jankovic2018risk}, and diseases typically result from complex interactions among clinical factors rather than from isolated individual factors \citep{guo2025explainable}.
Methods that provide interpretability for feature interactions therefore not only better represent the prediction process but also align with domain knowledge.

In response, a growing body of research aims to develop techniques that explicitly account for feature interactions. One notable line of work extends SHAP to incorporate interaction effects by evaluating subsets of features, thereby quantifying the importance of feature combinations \citep{tsai2023faith, sundararajan2020shapley}. 
 Some studies aim to develop intrinsic, interaction-based interpretability methods by extending generalized additive models (GAMs) to incorporate pairwise interaction terms \citep{chang2021node}, though these approaches are limited by the inherently additive nature of the GAM framework. Beyond these, \cite{janizek2021explaining} introduced Integrated Hessians to assess the importance of feature interactions, while \cite{guo2025explainable} developed a novel graph learning-based model to inherently identify complex interactions between patient attributes and their impact on length of stay  prediction. Furthermore, \cite{guo2025explainable} provides a comparative overview of individual-feature and feature-interaction-based interpretability methods, offering valuable insights for interested readers.

\subsection{Interpretability imperatives in time-series forecasting}

The interpretability of time-series data cannot be effectively addressed using post-hoc methods such as SHAP and related approaches, as these methods independently perturb individual time points, thereby disrupting the inherent sequential structure and leading to uninformative or even misleading explanations. Instead, we require a model intrinsically aligned with the sequential nature of the data—one that preserves temporal dependencies and explicitly characterizes cross-temporal feature interactions (i.e., interactions between different features across different time steps). Accordingly, we adopt a linear polynomial model for time-series forecasting:
$$
f(x) = \sum_{m=0}^{s} \; \sum_{1 \le t_1 \le \dots \le t_m \le T} \; \sum_{1 \le i_1 \le \dots \le i_m \le d} w^{(i_1,\dots,i_m)}_{t_1,\dots,t_m} \cdot \prod_{k=1}^{m} x^{(i_k)}_{t_k}.
$$
Table \ref{tab:polynomial_components} provides a detailed explanation of the components in this formula.

\begin{table}[htbp]
    \caption{Explanation of the polynomial formula components}
    \vspace{-0.5cm}
    \begin{center}
    	\scalebox{0.76}{
            \begin{tabular}{cl}
                \toprule
                \textbf{Component} & \textbf{Meaning} \\
                \midrule
                $s$ &  Total degree of the polynomial. \\
                $m$ & Degree of the current monomial ($0 \le m \le s$). \\
                $T$ & Total length of the time series. \\
                $d$ & Feature dimension at each time point. \\
                $t_k$ & Time index of the $k$-th variable ($1 \le t_k \le T$). \\
                $i_k$ & Feature index of the $k$-th variable ($1 \le i_k \le d$). \\
$1 \le t_1 \le \dots \le t_m \le T$ & Non-decreasing ordering of time indices.\\
$1 \le i_1 \le \dots \le i_m \le d$ & Non-decreasing ordering of feature indices at same time.\\
$w^{(i_1,\dots,i_m)}_{t_1,\dots,t_m}$ &  
Quantifies the importance of that spatio-temporal interaction pattern.\\
$\prod_{k=1}^{m} x^{(i_k)}_{t_k}$ & Monomial formed by multiplying $m$ variables;$x^{(i_k)}_{t_k}$ is the value of feature $i_k$ at time $t_k$. \\
                \bottomrule
            \end{tabular}}
    \end{center}
    \label{tab:polynomial_components}
\end{table}
  \vspace{-0.2cm}
  
We consider a special case when $s=2$:
$$
f(x) = w_0 
+ \sum_{t=1}^T \sum_{i=1}^d w^{(i)}_{t} x^{(i)}_t
+ \sum_{1 \le t_1 \le t_2 \le T} \sum_{1 \le i_1 \le i_2 \le d} w^{(i_1,i_2)}_{t_1,t_2} x^{(i_1)}_{t_1} x^{(i_2)}_{t_2},
$$
where $w_0$ is a constant bias term, $w^{(i)}_{t}$ captures the 
importance of feature $i$ at time $t$, and $w^{(i_1,i_2)}_{t_1,t_2}$  quantifies the importance of feature interaction between feature $i_1$ at time $t_1$ and feature $i_2$ at time $t_2$.  

The design of this polynomial model embodies a deep coupling between the model structure and its interpretability. The weight \textbf{$w^{(i)}_{t}$}  represents a ``time–feature'' effect, while $w^{(i_1, i_2)}_{t_1, t_2}$ quantifies a ``time–feature interaction'' effect, which is fundamentally different from post-hoc interpretability methods.

\subsection{Interpretable polynomial learning model}

Let $D=\{(x_i,y_i)\}_{i=1}^m$ be the set of samples with
$x_i\in\mathcal X\subseteq\mathbb R^d$, 
and $y_i\in[-M,M]$
for some positive number $M$.  
Our purpose is
\begin{equation}\label{KRR2}
         \arg\min_{f\in\mathcal
          P_s^d}\left\{\frac1m\sum_{i=1}^m\phi(f(x_i),y_i)\right\}
\end{equation}
for arbitrary $s$.
 Since the dimension of $\mathcal P_s^d$
is $ n=\left(^{s+d}_{\ s}\right)$, if we can select
$\{\eta_i\}_{i=1}^n\subset\mathbb R^d$ such that $\{(1+\eta_i\cdot
x)^s\}_{i=1}^n$ is a linear independent system, then
\begin{equation}\label{hypotheisis n space}
        \mathcal P_s^d=\left\{\sum_{i=1}^nc_i(1+\eta_i\cdot
        x)^s:c_i\in\mathbb R\right\}=:\mathcal H_{\eta,n},
\end{equation}
 where $\mathcal H_{\eta,n}$ is the finite-basis linear space constructed for solving (1).
In this way, (\ref{KRR2}) can be converted to
\begin{equation}\label{new model}
        \arg\min_{f\in\mathcal H_{\eta,n}} \left\{\frac1m\sum_{i=1}^m\phi(f(x_i),y_i)\right\}.
\end{equation}

To verify the feasibility of   (\ref{new model}), there are two
things we should do. The one is to give a selection strategy of
$\{\eta_i\}_{i=1}^n$ such that (\ref{hypotheisis n space}) holds and
 the other is to guarantee the non-singularity of the matrix
$A_{m,n}:=((1+x_i\cdot \eta_j)^s)_{i,j=1}^{m,n}$.


We first present the selection strategy for $\{\eta_i\}_{i=1}^n$ based on the conceptions of Haar space and
fundamental system \citep{wendland2004scattered}. 
Following \citep{wendland2004scattered}, if a set of points $\{\eta_i\}_{i=1}^n$ in $\mathbb R^d$ can be chosen such that $\mathcal H_{\eta,n}$ forms an $(n+1)$-dimensional Haar space, then all the aforementioned issues can be resolved.
However, for $d \geq 2$, there does not exist a Haar space on $\mathbb B^d$ with dimension $N \geq 2$ \citep{lin2018fast}. In this context, we employ the fundamental system associated with the polynomial kernel $K_s$, as introduced in \citep{lin2018fast}.





\begin{definition}\label{FUNDAMENTAL SYSTEM}
Let $\zeta:=\{\zeta_i\}_{i=1}^n\subset\mathbb R^d$. $\zeta$ is
called a $K_s$-fundamental system if
$
          \mbox{dim}{\mathcal
            H_{\zeta,n}}=\left(_{\ s}^{s+d}\right).
$
\end{definition}

From the above definition, it is easy to see that arbitrary
$K_s$-fundamental system implies (\ref{hypotheisis n space}). 
Moreover, \cite[Prop.1]{lin2018fast} demonstrates that almost all $n=\left(_{\
s}^{d+s}\right)$ points  constitutes a $K_s$-fundamental system.


Based on \citep[Prop.1]{zeng2019fast}, we can design   simple
strategies to choose the centers $\{\eta_j\}_{j=1}^n$. In
particular, $\{\eta_j\}_{j=1}^n$ can be selected
randomly independently and
identically according to the uniform distribution, since the uniform
distribution is continuous with respect to Lebesgue measure
\citep{bass2005}.
 Then   there almost surely holds
$$
        \mathcal P_s^d=\left\{\sum_{i=1}^nc_i(1+\eta_i\cdot
        x)^s:c_i\in\mathbb R\right\}.
$$

According to \citep{lin2018fast}, since $n$ is typically much smaller than $m$, the matrix $A_{m,n}$ is almost surely nonsingular with probability close to one, which establishes the feasibility of the model in (\ref{new model}).

\subsection{Interpretable polynomial learning algorithm for time series}

Building on the IPL model introduced in the previous section, we now present its adaptation for time‑series forecasting.  Before doing so, we first describe the setting of time-series data. Unlike independent and identically distributed (i.i.d.) samples, observations in time series are inherently temporally dependent. Specifically, temporal dependence exists not only between inputs and outputs, but also among inputs at different time steps and among outputs across time. Therefore, in addition to incorporating individual input features and their interactions—as formulated in \eqref{hypotheisis n space}—the time‑series IPL algorithm must flexibly embed temporal information into the model. In particular, the model input is extended to include lagged variables, namely past observations of both the input features and the target labels. To align with the requirements of early-warning applications, where users typically focus on the most influential features, we introduce an interpretability threshold that selects only the most relevant terms, thereby yielding a sparse yet interpretable model.

According to Algorithm~\ref{alg:IPL}, given a dataset, IPL requires only the specification of the polynomial degree $s$, the numbers of lagged input variables ($L_x$) and target labels ($L_y$), and an interpretability threshold $I$. Based on these settings, the algorithm produces an interpretable estimator that simultaneously quantifies the contributions of individual features and their interactions to each prediction. The proposed algorithm offers the following advantages:


\begin{itemize}
        \item \textit{Native support for time-series forecasting:} The feature-level interpretability of IPL is intrinsically coupled with its model architecture.  It not only accommodates the temporal dependencies inherent in time-series data, but also allows for the flexible incorporation of lagged variables as inputs. This design is consistent with widely adopted practices in time-series forecasting \citep{bertani2025joint, hen2025spatio}.

            \item \textit{Flexible accuracy–interpretability trade-off:}  IPL achieves a balance between accuracy and interpretability simply by adjusting the polynomial degree.
            Unlike deep learning models that require extensive hyperparameter tuning and domain expertise, IPL’s simplicity makes it accessible to non-specialists and thereby facilitates practical use in early warning.

    \item \textit{Low computational overhead for early-warning applications:} IPL employs ADMM for optimization. For a wide range of loss functions, efficient ADMM-based solution strategies are already available, ensuring the model’s scalability and facilitating its implementation in real-time early-warning applications.

\end{itemize}

\begin{algorithm}[!t]
\small
\caption{Interpretable Polynomial Learning for Time Series}
\label{alg:IPL}
 \textbf{1. Input}: A sample set $D=\{(x_i, y_i)\}_{i=1}^T$, where $T$ denotes the final data collection timestamp, $x_i \in \mathbb{R}^d$, the polynomial kernel $K_s(x, x') = (1 + x \cdot x')^s$, the polynomial degree $s \in \mathbb{N}$, time lag parameters $L_x, L_y$, and  an interpretability threshold $I$.
 
 \textbf{2. Determine kernel centers:} Set $n = \binom{s+d}{s}$ (the dimension of polynomial space of degree $s$ in $d$ variables) and take $\{n_j\}_{j=1}^n$ with $n_j = x_j$ as the $K_s$ fundamental system.

 \textbf{3. Construct new time series}: Define $\tilde x_i = (x_i, x_{i-1},\dots,x_{i-L_x}, y_{i-1},\dots,y_{i-L_y}) \in \mathbb{R}^{d+L_x+L_y}$, $0 \leq L_x<i, 0 \leq L_y<i$,
 which yields the expanded sample set $\tilde D=\{(\tilde x_i, y_i)\}_{i=1}^T$. Form the kernel matrix 
$
A := \big(K_s(\tilde x_i, n_j)\big)_{i=1,j=1}^{T,n}.
$
 
 \textbf{4. Solve objective function (\ref{new model}) by ADMM}:
Reformulate (\ref{new model}) as the following 
optimization problem via introducing another variable \( v \),
\begin{equation}\label{unconstrained_problem}
\min_{u \in \mathbb{R}^n, v \in \mathbb{R}^T} f(v) \quad \text{s.t.} \quad Au - v = 0,
\end{equation}\label{KRR2}
where \( f(v) := \frac{1}{T} \sum_{i=1}^{T} \phi(v_i,y_i) \). The augmented Lagrangian function of (\ref{unconstrained_problem}) is defined by
\[
\mathcal{L}_\beta (u, v, w) = f(v) + \langle w, Au - v \rangle + \frac{\beta}{2} \| Au - v \|_2^2,
\]

where \( w \in \mathbb{R}^T \) is a multiplier variable, \( \beta > 0 \) is the augmented Lagrangian parameter. Based on \(\mathcal{L}_\beta\), the ADMM algorithm for (\ref{new model}) is as follows: given an initialization \( u^0, v^0, w^0 \), parameters \( \alpha > 0, \beta > 0 \), for \( k = 0, 1, \ldots \),
\begin{align}
u^{k+1} &= \arg \min_{u \in \mathbb{R}^n} \left\{ \mathcal{L}_\beta (u, v^k, w^k) + \frac{\alpha}{2} \| u - u^k \|_2^2 \right\}, \tag{5} \\
v^{k+1} &= \arg \min_{v \in \mathbb{R}^T} \mathcal{L}_\beta (u^{k+1}, v, w^k), \tag{6} \\
w^{k+1} &= w^k + \beta (Au^{k+1} - v^{k+1}). \tag{7}
\end{align}
Apply the proximal update strategy to compute $u^{k+1}$  in (5),  update $v^{k+1}$ according to the specific loss function $\phi$, and finally obtain the global minimizer of (3): $f_{\tilde{D},s}(\tilde{x}) = \sum_{j=1}^n u_j^* K_s(n_j, \cdot)$.

\textbf{5. Extract feature importance}:
Express the predictor $f_{\tilde{D},s}(\cdot)$ as an explicit polynomial expansion over the $(d+L_x+L_y)$-dimensional input $\tilde x$:
\[
f_{\tilde{D},s}(\tilde x) = \sum_{|\boldsymbol{\alpha}| \le s} \omega_{\boldsymbol{\alpha}} \; \tilde x_1^{\alpha_1} \cdots \tilde x_{d+L_x+L_y}^{\alpha_{d+L_x+L_y}},
\]
where $\boldsymbol{\alpha} = (\alpha_1, \dots, \alpha_{d+L_x+L_y})$ is a multi-index with $\alpha_j \in \mathbb{N}_0$ denoting the exponent of the $j$-th variable, and
$|\boldsymbol{\alpha}| := \sum_{j=1}^{d+L_x+L_y} \alpha_j$ denotes the order of the multi-index (while $|\cdot|$ applied to a scalar represents the absolute value).
The coefficient $\omega_{\boldsymbol{\alpha}}$ quantifies the contribution of the corresponding monomial term to the prediction.

\textbf{6. Select features and coefficients}: 
Collect the surviving multi‑indices $\boldsymbol{\alpha}$ and their corresponding coefficients $\omega_{\boldsymbol{\alpha}}$ into a sparse set 
$
\Omega_I := \bigl\{ (\boldsymbol{\alpha},\,\omega_{\boldsymbol{\alpha}}) \mid |\boldsymbol{\alpha}| \le s,\; |\omega_{\boldsymbol{\alpha}}| \ge I \bigr\}.
$

\textbf{7. Output}: The interpretable predictor $
f_{\tilde{D},s}^{I}(\tilde x) = \sum_{(\boldsymbol{\alpha},\omega_{\boldsymbol{\alpha}}) \in \Omega_I} \omega_{\boldsymbol{\alpha}} \; \tilde x_1^{\alpha_1} \cdots \tilde x_{d+dL_x+L_y}^{\alpha_{d+dL_x+L_y}},
$ and its sparse coefficient set $\Omega_I$.

\end{algorithm}

\begin{remark}
For Step~4 of the IPL algorithm, the update of $u^{k+1}$ can be expressed analytically as $$u^{k+1} = \left( \beta A^\top A + \alpha I_n \right)^{-1} \left[ \alpha u^k + \beta A^\top v^k - A^\top w^k \right],$$ where $I_n$ is the identity matrix of size $n$.
For the squared loss, an analytical solution to~\eqref{new model} is given by \citep{lin2018fast}:
$
f_{\tilde{D},s}(\cdot) = \sum_{j=1}^n c_j K_s(n_j, \cdot),
$
with $\mathbf{c} = (c_1, \dots, c_n)^{\top}$ given by $\mathbf{c} = \mathrm{pinv}(A)\mathbf{y}$. Here, $\mathrm{pinv}(\cdot)$ denotes the pseudo-inverse operator; while for the hinge loss, a fast iterative solver is available in \citep{zeng2019fast}. 

If $\phi$ is differentiable, the $v$-update decomposes into $T$ independent scalar equations:
$$
\frac{1}{T} \phi'(v_i, y_i) + \beta (v_i - b_i) = 0,
$$
where $b_i = (A u^{k+1})_i + w_i^k / \beta$. For convex loss functions (such as squared, logistic, or hinge loss), each equation corresponds to minimizing a one‑dimensional convex function. It can therefore be solved efficiently by Newton’s method: for each $i$,
$$
v_i^{(t+1)} = v_i^{(t)} - \frac{\frac{1}{T} \phi'(v_i^{(t)}, y_i) + \beta (v_i^{(t)} - b_i)}{\frac{1}{T} \phi''(v_i^{(t)}, y_i) + \beta},
$$
which typically converges in a few iterations. 

Specifically, for the logistic loss $\phi(v, y) = \log(1 + e^{-y v})$ with $y \in \{-1, 1\}$, the Newton iteration becomes
$$
v_i^{(t+1)} = v_i^{(t)} - 
\frac{\frac{y_i}{T} \bigl[ \sigma(y_i v_i^{(t)}) - 1 \bigr] + \beta (v_i^{(t)} - b_i)}
{\frac{1}{T} \sigma(y_i v_i^{(t)}) \bigl[ 1 - \sigma(y_i v_i^{(t)}) \bigr] + \beta},
$$
where $\sigma(z) = 1/(1+e^{-z})$. 

\end{remark}

\section{Simulated data analysis under the squared loss}

In this section, we conduct three sets of experiments to evaluate the effectiveness of the proposed IPL under the squared loss function. First, we investigate the importance of time-dependent target variables in time-series analysis. Second, we demonstrate the advantages of IPL in terms of feature interpretability, prediction accuracy, and computational efficiency. Finally, we perform a perturbation analysis by disturbing the most important features identified by each interpretability method, in order to evaluate which method most accurately identifies the features critical for accurate predictions.

The interpretability methods used in this study are divided into two categories:
(1) Post-hoc interpretability methods: SHAP and LIME. For SHAP, feature contributions are computed using a deep neural network (DNN), whereas in LIME, kernel ridge regression (KRR) serves as the local surrogate model.
(2) Ante interpretability method: ARIMAX. Detailed experimental configurations for SHAP, LIME, and ARIMAX are provided in the Appendix. Unless otherwise stated, the same settings for these four methods are applied in all subsequent experiments. In particular, the polynomial degree $s$ for IPL is set to 2 based on its optimal performance in our numerical validation.


We generate time-series data comprising training samples $\{(x_t, y_t)\}_{t=1}^{T}$ and testing samples $\{(x_{t'}^*, y_{t'}^*)\}_{t'=1}^{T'}$. The covariates $x_t$ and $x_{t'}^*$ are independently and identically drawn from the 5-dimensional hypercube $[0,1]^5$ following a uniform distribution. The temporal evolution of the series is described by objective function $
y_{t+1} = \cos(y_t) \sin(y_{t-1}) + f(x^t) + \epsilon^t,
$
$
y_{t'+1}^* = \cos(y_{t'}^*) \sin(y_{t'-1}^*) + f(x_{t'}^*),
$
 the function $f(\cdot)$ is defined as follows with $\epsilon^t\sim \mathcal{N}(0, 0.1^2)$
$$
f(\cdot) =  
(1+x_1^t+2x_2^t+3x_3^t+4x_4^t+5x_5^t+6x_1^tx_2^t-7x_3^tx_4^t)/15.
$$

The sample sizes are set to $T = 4,000$ for training and $T' = 1,000$ for testing. All experimental results are averaged over 10 independent trials. Except for the LIME-related experiments, which were performed on a GPU, all other experiments were conducted using Python 3.7 on a PC with an Intel Core i5 2GHz processor. The scripts for reproducing these experiments are available at \url{https://github.com/Ariesoomoon/IPL_TS_experiments}.


\subsection{Importance of temporal dependencies in time series}

\begin{figure}[H]
	\centering
	\vspace{0.1in}
	\includegraphics[scale=0.38]{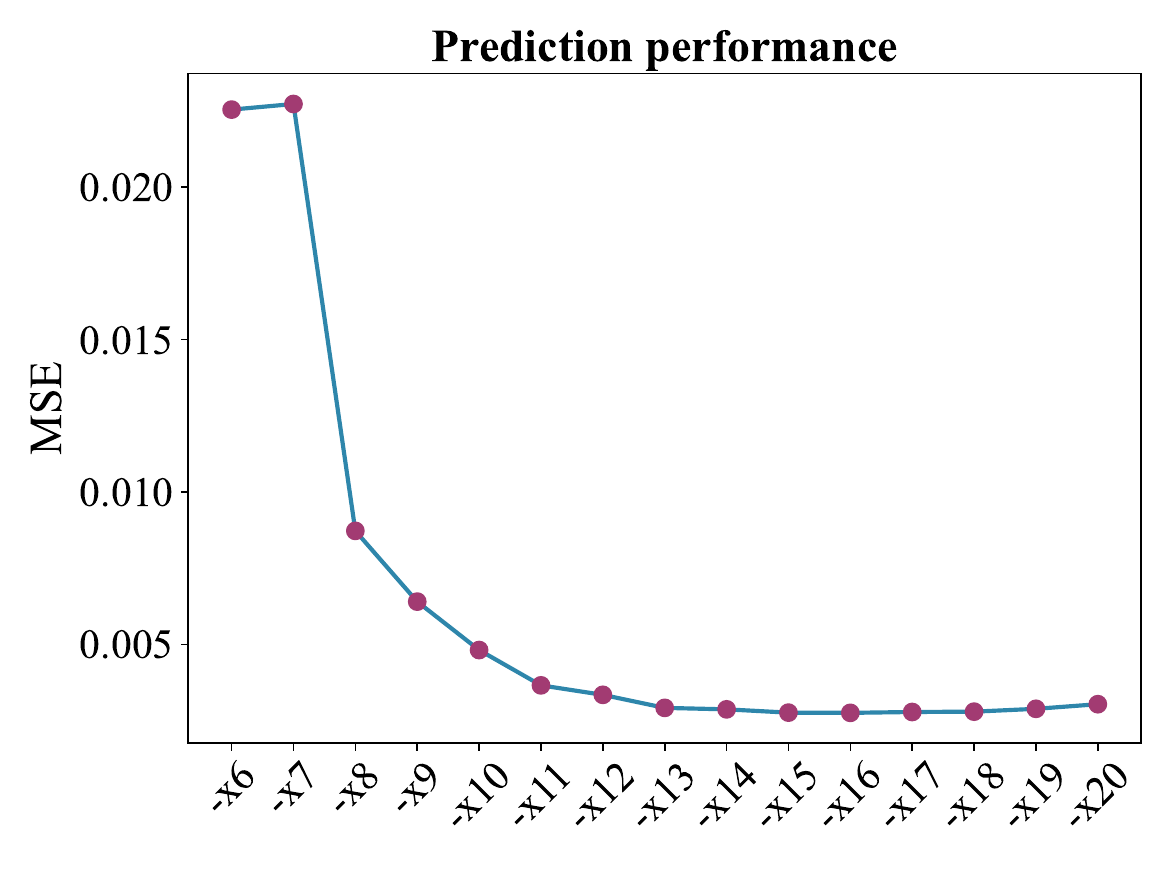}
\caption{Importance of time-related target variables (``-x6'' denotes a 6-dimensional input formed by combining the original features with one lagged target variable, ``-x7'' with two lagged target variables, and so on).}
	\label{Fig:sim_data_continous_x6tox20_average}  
	\vspace{-0.3cm}
\end{figure}

Figure \ref{Fig:sim_data_continous_x6tox20_average} illustrates how integrating lagged target variables affects regression performance. Here, ``-x6'' denotes a 6-dimensional input formed by combining the original features with the target variable from the previous time lag, ``-x7''  represents the original input combined with the target variable from the two preceding lags, and so forth.
The results demonstrate a consistent improvement in prediction accuracy as more lagged variables are included, highlighting the value of incorporating historical target information in time-series forecasting. However, starting from $x_{13}$, we observe that the prediction accuracy stabilizes.  This may be attributed to the fact that the negative impact of introducing redundant features begins to neutralize the prediction value gained from incorporating additional lagged target variables.

\vspace{-0.3cm} 

\subsection{Advantages of IPL in providing feature interpretability}

This experiment compares IPL with SHAP, LIME, and ARIMAX to demonstrate the advantages of IPL.  We introduce two lagged target variables $x_7, x_8$, and an irrelevant input variable $x_6$ to enhance credibility in examining each method's ability to exclude it from consideration.

\begin{figure}[H]
	\centering
	\vspace{0.1in}
	\setlength{\subfigcapskip}{-0.5em}
	\subfigure[IPL (Ante-hoc)]{\includegraphics[scale=0.25]{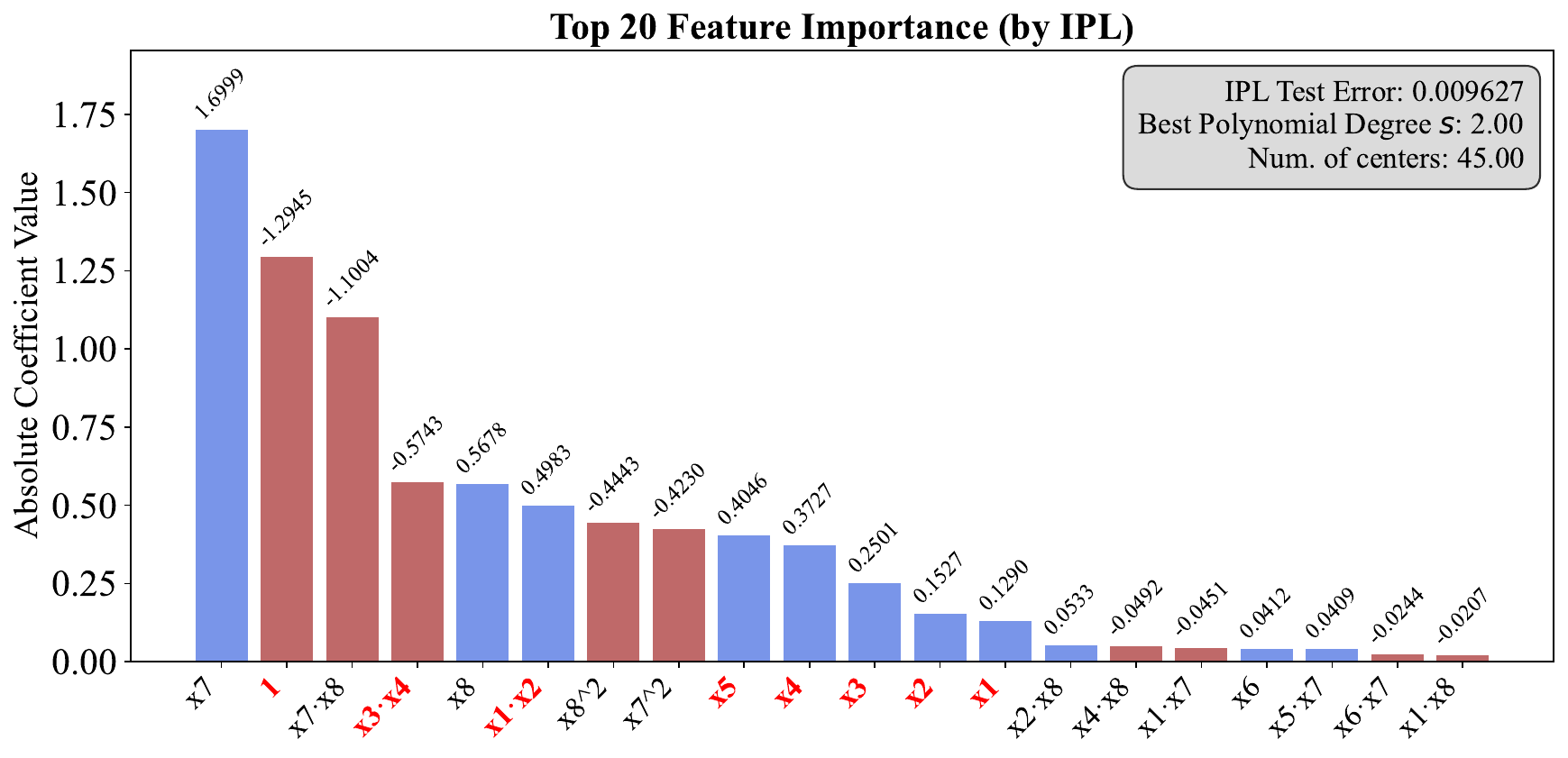}
		\label{subfig: fpc_feature_importance_x1tox8_mean}} 
                    	\setlength{\subfigcapskip}{-0.5em}
	\subfigure[ARIMAX (Ante-hoc)]{\includegraphics[scale=0.25]{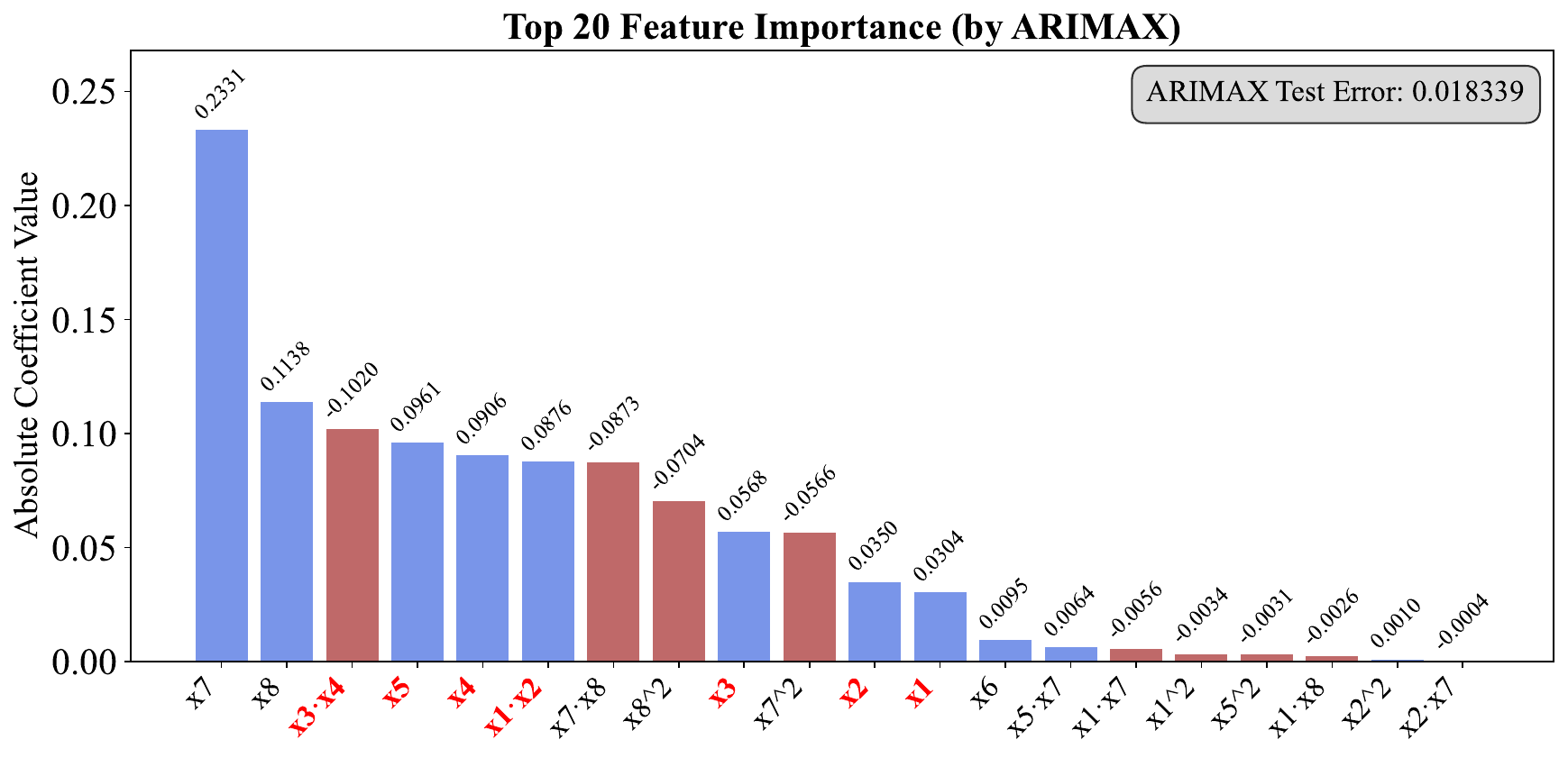}
		\label{subfig: arimax_feature_importance_from_saved_data}} 
	\setlength{\subfigcapskip}{-0.5em}
	\subfigure[DNN+SHAP (Post-hoc)]{\includegraphics[scale=0.25]{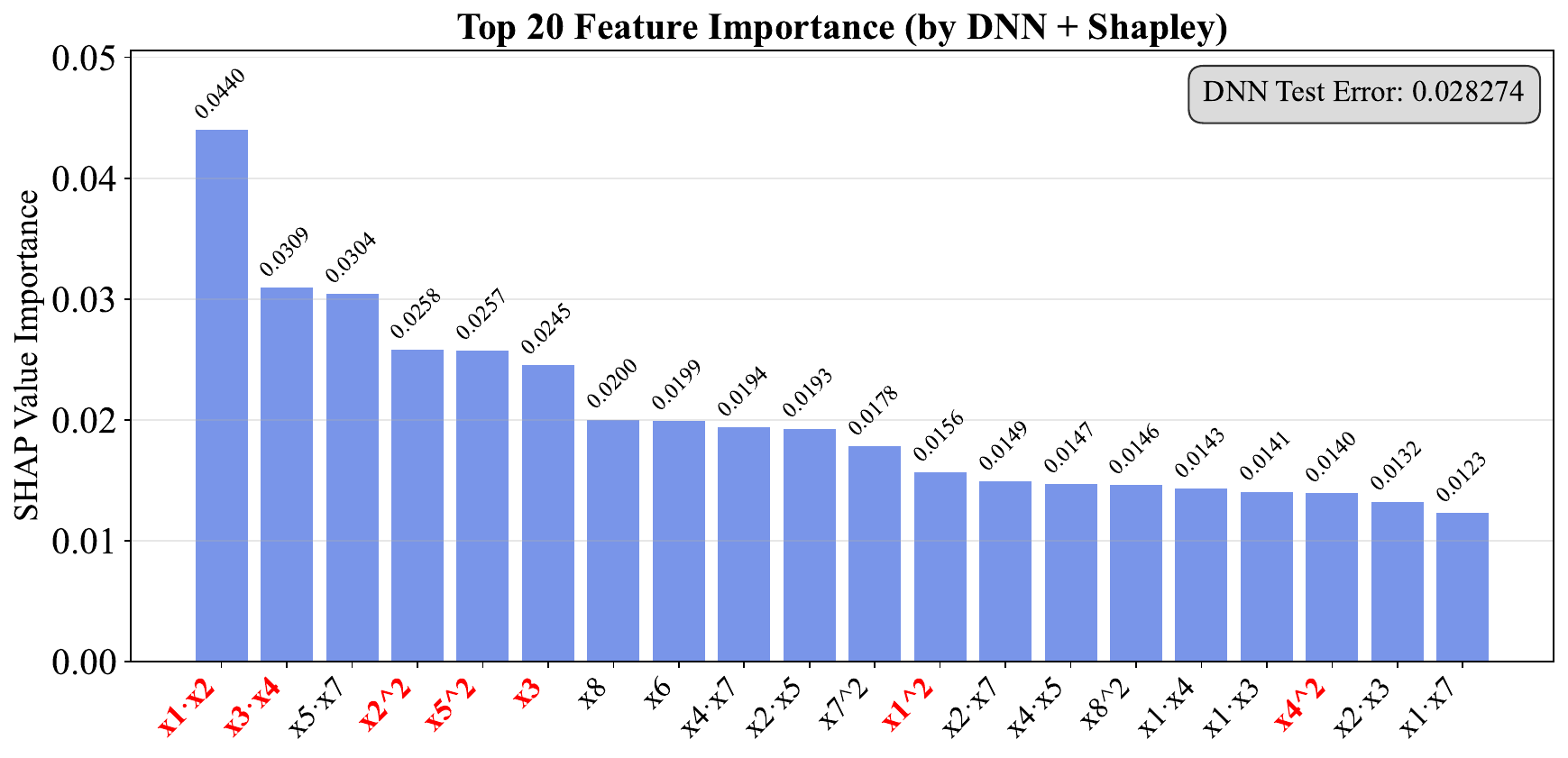}
		\label{subfig: shap_feature_importance_x1tox8_mean}} 
        	\setlength{\subfigcapskip}{-0.5em}
	\subfigure[KRR+LIME (Post-hoc)]{\includegraphics[scale=0.25]{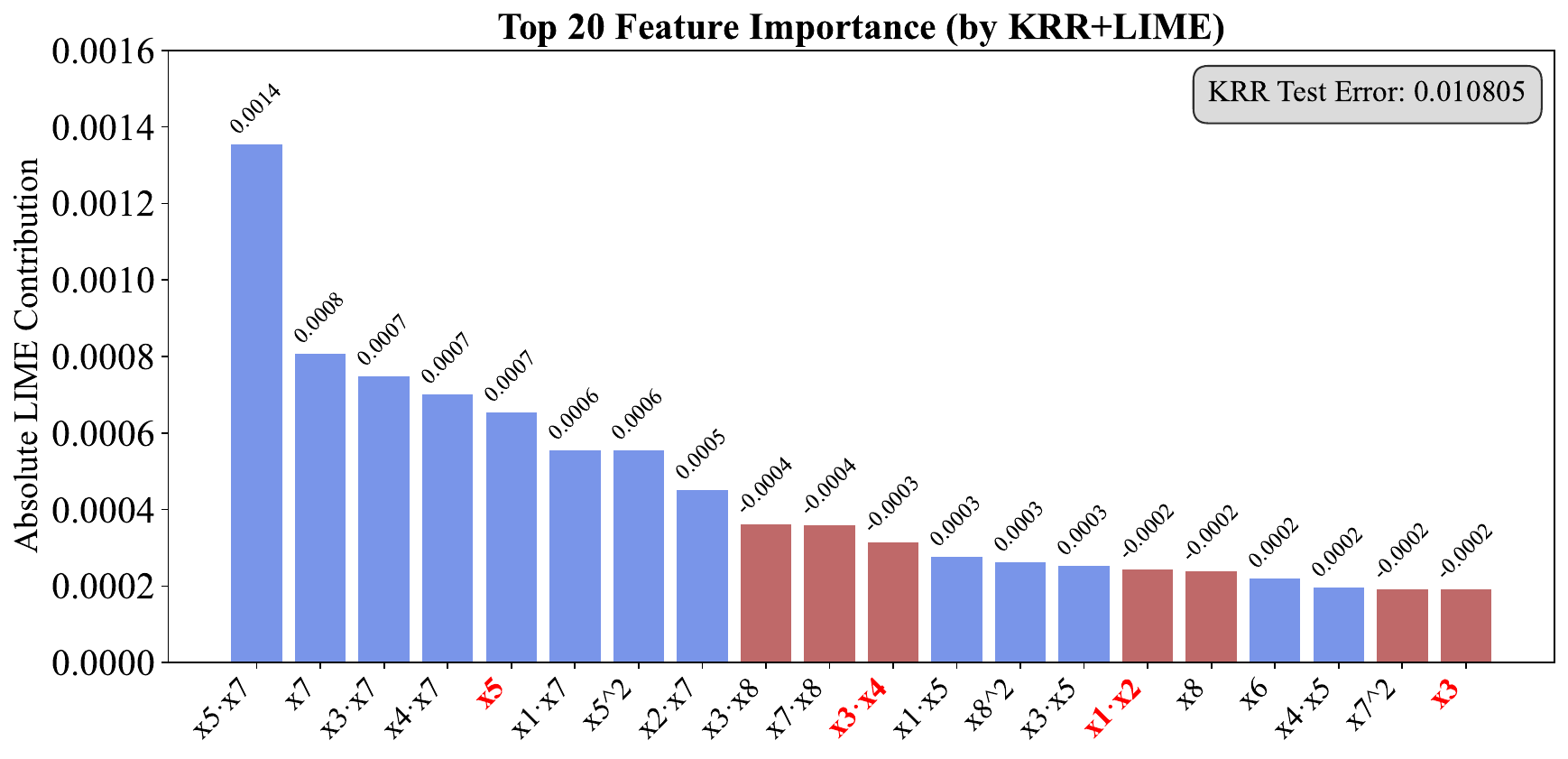}
		\label{subfig: top_features_LIME_importancee_x1tox8_mean}} 
            	\caption{Top important features identified by different interpretability methods (note: features shown on the horizontal axis that are present in the objective function are highlighted in red; red and blue bars in the chart represent negative and positive values, respectively).}
	\label{fig: Top Important Feature x1tox8}
\end{figure}

Figure \ref{fig: Top Important Feature x1tox8} presents the top 20 important features identified by different interpretability methods, with IPL's prediction accuracy intentionally maintained at the lowest level to ensure a  robust comparison of their feature interpretability capabilities. 
Three key observations can be drawn:
(1) Only IPL achieves feature importance rankings and magnitudes that are fully consistent with the objective function. While ARIMAX also performs well in this regard, its training process is  slower in practice, as it requires identifying the optimal $(p, d, q)$ orders through grid search.
(2) IPL effectively deprioritizes the irrelevant feature $x_6$ (ranking it $17^{\text{th}}$). Although LIME demonstrates a similar ability to identify irrelevant variables, it demands significantly more computational time and memory. In fact, the LIME results in this experiment were obtained using GPU acceleration, with its runtime being approximately 1000 times longer than that of IPL executed on a CPU.  (3) All methods, particularly intrinsically interpretable ones such as IPL and ARIMAX, successfully identify the importance of the incorporated lagged target variable.

\vspace{-0.5cm}

\begin{table}[H]
	\renewcommand\arraystretch{1.15}
	\caption{Performance comparison of interpretability methods under different evaluation metrics}
	\label{tab:methods_comparison_multi}
    	\begin{center}
\vspace{-0.5cm} 
	\scalebox{0.65}{
	\begin{tabular}{lcccccccc}
		\hline
        		\cline{2-7}
		\multicolumn{1}{l}{ \multirow{2}*{\textbf{Metrics}
		} } 	& \multicolumn{4}{c}{\makecell[c]{Original input with three added input variables \\ $x_6, x_7, x_8$ (Dimension = 8)}} & \multicolumn{4}{c}{\makecell[c]{Original input only\\(Dimension = 5)}} \\
		\multicolumn{1}{c}{} & IPL & SHAP & LIME& ARIMAX & IPL & SHAP & LIME& ARIMAX \\
		\hline
		\multicolumn{7}{l}{\textbf{Feature interpretability}} \\
		\textit{Feature overlap ratio} & \textcolor{blue}{0.7143} & 0.4286 & 0.1429  &  0.7143& \textcolor{blue}{1.0000} & 0.4286 & 0.7143&   0.8571   \\
		\textit{Ranking similarity}  & \textcolor{blue}{1.0000} & 0.6699 & 0.7881 &  0.8929   & \textcolor{blue}{1.0000} & 0.8571 & 0.6847& 0.4324   \\
		\textit{Value similarity}  & \textcolor{blue}{0.9979} & 0.1734 & 0.4723&   0.9835 & \textcolor{blue}{0.9992} & 0.2759 & 0.2831& 0.8487    \\
		\hline
		\multicolumn{7}{l}{\textbf{Prediction accuracy (MSE) on top 10 features}} \\
		Linear regression (LR) & \textcolor{blue}{0.0064} & 0.0073 & 0.0096&  0.0064  & 0.0159 & 0.0168 & 0.0165&  \textcolor{blue}{0.0158}  \\
        \hline
		\textbf{Efficiency (runing time/trial)}
        & \textcolor{blue}{1.81 \textit{sec}} & 32.45 \textit{sec} & 32.50 \textit{min}& 20.57 \textit{sec}& \textcolor{blue}{1.47 \textit{sec}} & 20.55 \textit{sec} & 17.03 \textit{min}&  5.56 \textit{sec} \\
		\hline
	\end{tabular}
	}
    \vspace{3pt} 
		\footnotesize
		\begin{tabular}{@{}l@{}}
			\multicolumn{1}{@{}p{0.97\linewidth}@{}}{
				\scriptsize
				\textbf{Note}: 
				The best results for each metric are highlighted in color. 
\textit{Feature overlap ratio} refers to the proportion of features from the target function that are also included in the method's top-ten feature set.
\textit{Ranking similarity}  is defined as
$\rho = 1 - ( 6\sum_{i=1}^n d_i^2 ) / ( n(n^2-1) )$
where $d_i = R_\text{method}(i) - R_\text{target}(i)$ denotes the rank difference of the $i$-th feature between the method’s ranking $R_\text{method}$ and the ground truth ranking $R_\text{target}$.
\textit{Value similarity} is defined as
$
\frac{\mathbf{a}\cdot\mathbf{b}}{\|\mathbf{a}\|\|\mathbf{b}\|}
= (\sum_{i=1}^n a_i b_i) / (\sqrt{\sum_{i=1}^n a_i^2}\,\sqrt{\sum_{i=1}^n b_i^2}),
$
where $\mathbf{v}=(v_1,\dots,v_n)$ 
denotes the vector across $n$ target features.
 ``\textit{sec}'' denotes seconds and ``\textit{min}'' denotes minutes.
			}
		\end{tabular}
        	\end{center}
\end{table}
\vspace{-1cm}

\begin{figure}[H]
	\centering
	\vspace{0.1in}
	\includegraphics[scale=0.42]{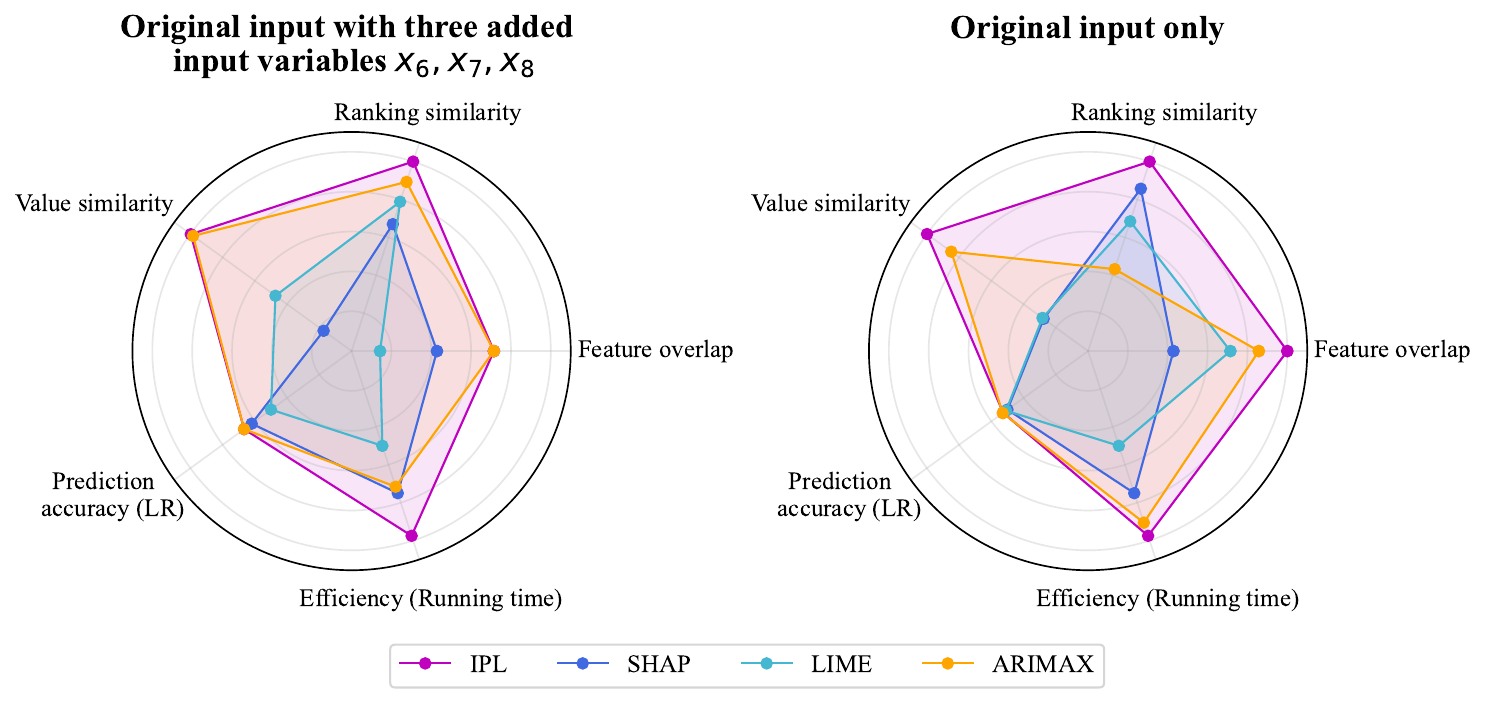}
\caption{Radar charts comparing the performance of interpretability methods (closer proximity to the outer boundary indicates stronger capability in that dimension).}
	\label{Fig:radar_charts_comparison}  
	\vspace{-0.3cm}
\end{figure}

We next conduct a comprehensive comparison of the different methods in terms of three key dimensions: feature interpretability, prediction accuracy, and computational efficiency. As shown in Table \ref{tab:methods_comparison_multi}, IPL consistently outperforms both SHAP and LIME in all evaluated aspects. Specifically, it provides more faithful interpretations by identifying features that closely align with the true feature importance defined in the objective function; it delivers the best prediction accuracy when using its identified top-ten important features; and it requires the least running time during the testing phase. To visually summarize the strengths of IPL, we present a radar chart for the results in Table \ref{tab:methods_comparison_multi}. As shown in Figure \ref{Fig:radar_charts_comparison}, IPL—represented by the red shaded region—occupies the largest area, clearly demonstrating its overall advantage in simultaneously achieving high interpretability, accuracy, and efficiency.

\subsection{Perturbation analysis}

We employed a perturbation analysis based on the same principle as the widely used permutation feature importance method \citep{guo2024measuring} to validate the feature importance identified by different interpretability methods. A significant degradation in predictive performance after perturbation indicates that the feature is truly important, thereby validating the effectiveness of the corresponding interpretability method.
Specifically, for each method’s most important feature, we add Gaussian noise $\epsilon \sim \mathcal{N}(0, \sigma^2)$, where $\sigma$ denotes the standard deviation of that feature, to perturb it while keeping all other features unchanged. We then compare the test MSE before and after perturbation over 10 trials, with linear regression consistently used as the estimator to ensure a fair comparison.


\begin{figure}[H]
	\centering
	\vspace{0.1in}
	\includegraphics[scale=0.35]{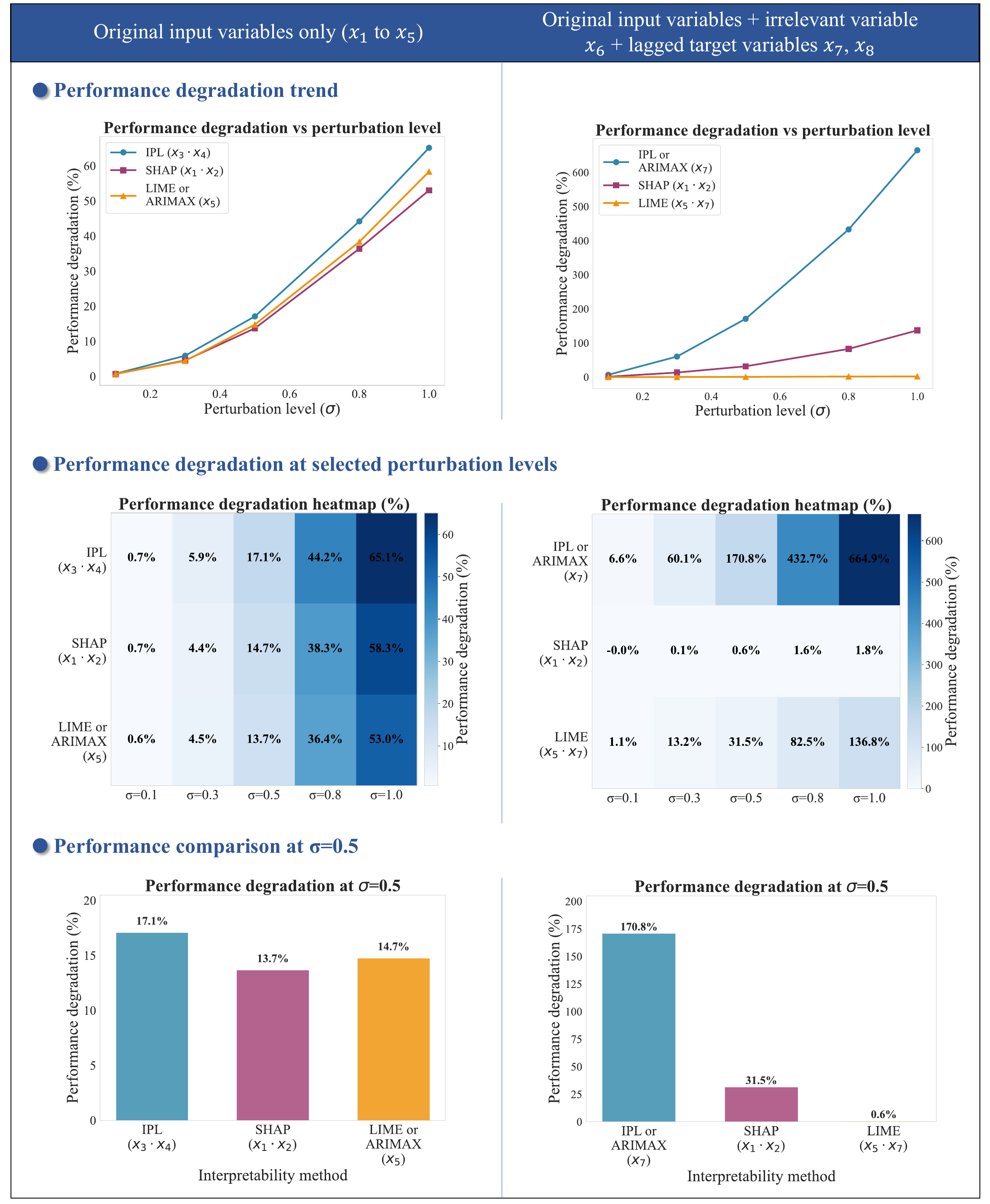}
\caption{Perturbation analysis (the most important feature selected by each method is indicated in parentheses).}
	\label{fig: perturbation_analysis}
\end{figure}
\vspace{-0.3cm}

The results are shown in Figure \ref{fig: perturbation_analysis}, where the left shows the perturbation analysis on the original data and the right shows the perturbation analysis when one irrelevant input variable ($x_6$) and two lagged target variables ($x_7$ and $x_8$) are added as input variables. To visually highlight both the trend and magnitude of performance changes, the results are presented using a combination of curves and a heatmap.
We see that perturbing the most important feature identified by IPL leads to the most significant performance degradation across all perturbation levels, demonstrating IPL’s advantage in feature interpretability. This advantage becomes even more pronounced after incorporating lagged target variables.
We further compare the three methods using a bar chart at the perturbation level of $\alpha = 0.5$. The results show that, after adding the lagged target variables, IPL exhibits a maximum performance degradation of 170.8\%, significantly higher than the 31.5\% observed for SHAP and the 0.6\% for LIME.

\section{Bitcoin historical price data analysis  under the hinge loss}

In financial time-series analysis, 
predicting the price movement patterns across different forecasting horizons is important, as it directly impacts trading decisions and profitability.
In this study, we utilized Bitcoin historical price data spanning from January 1, 2024 to present\footnote{https://www.kaggle.com/datasets/mczielinski/bitcoin-historical-data/data}. The dataset includes attributes such as opening price, highest price, lowest price, trading volume, and closing price. We used the opening price, highest price, lowest price, and trading volume as inputs, with the closing price as the output. 
To ensure computational efficiency while maintaining temporal continuity, we systematically sampled approximately 5,000 data points at regular intervals from the original high-frequency dataset, preserving the chronological order of observations.


\begin{figure}[H]
	\centering
	\vspace{0.1in}
	\includegraphics[scale=0.35]{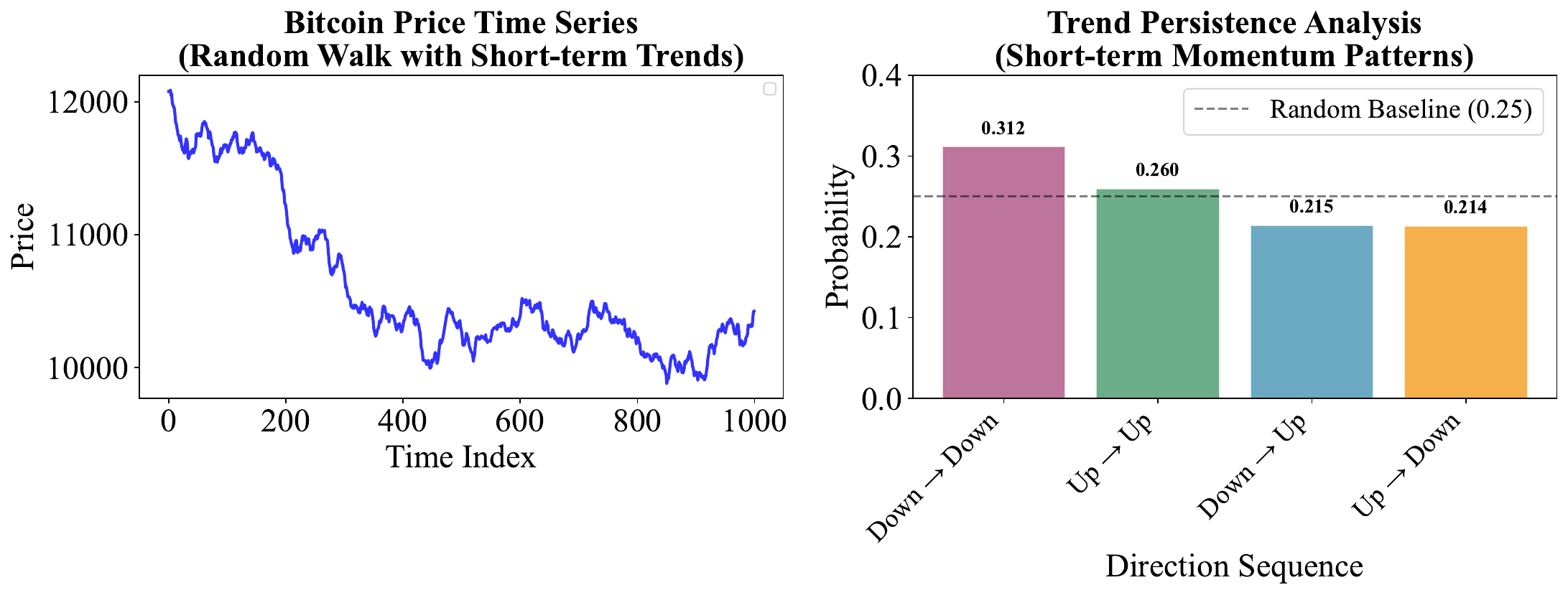}
            	\caption{Analysis of Bitcoin price dynamics}
	\label{Fig:price_characteristics_analysis}  
	\vspace{-0.3cm}
\end{figure}

We begin by examining the directional patterns of price changes.  As shown in 
Figure \ref{Fig:price_characteristics_analysis}, where  ``Up $\rightarrow$ Up'' describes cases where an upward trend at one time point is followed by another upward trend at the next time point. The analysis shows that both ``Up $\rightarrow$ Up'' and ``Down $\rightarrow$ Down'' cases occur with probabilities significantly above the  random baseline (0.25), suggesting the presence of short-term momentum—that is, prices tend to continue rising after prior increases and to keep falling after prior declines. Based on this observation, we formulate two binary classification tasks to model price movements over different forecasting horizons and employ the hinge loss function for model training. The target variables, labeled as \texttt{Target\_$k$period}, are defined for $k \in \{3, 5\}$, where $k$ denotes the prediction horizon. Each target encodes the price direction using a binary scheme: 1 indicates a price increase, while $-1$ denotes a price decrease after $k$ temporal units. The target variables are mathematically defined as:
\begin{equation}
\text{Target\_$k$period} = 
\begin{cases} 
1 & \text{if } P_{t+k} > P_t \\
-1 & \text{otherwise}
\end{cases}
\end{equation}
where $P_t$ represents the closing price at time $t$.



\subsection{Importance of temporal dependencies in time series}

\begin{figure}[H]
	\centering
	\vspace{0.1in}
	\setlength{\subfigcapskip}{-0.5em}
	\subfigure[\texttt{Target\_$3$period}]{\includegraphics[scale=0.3]{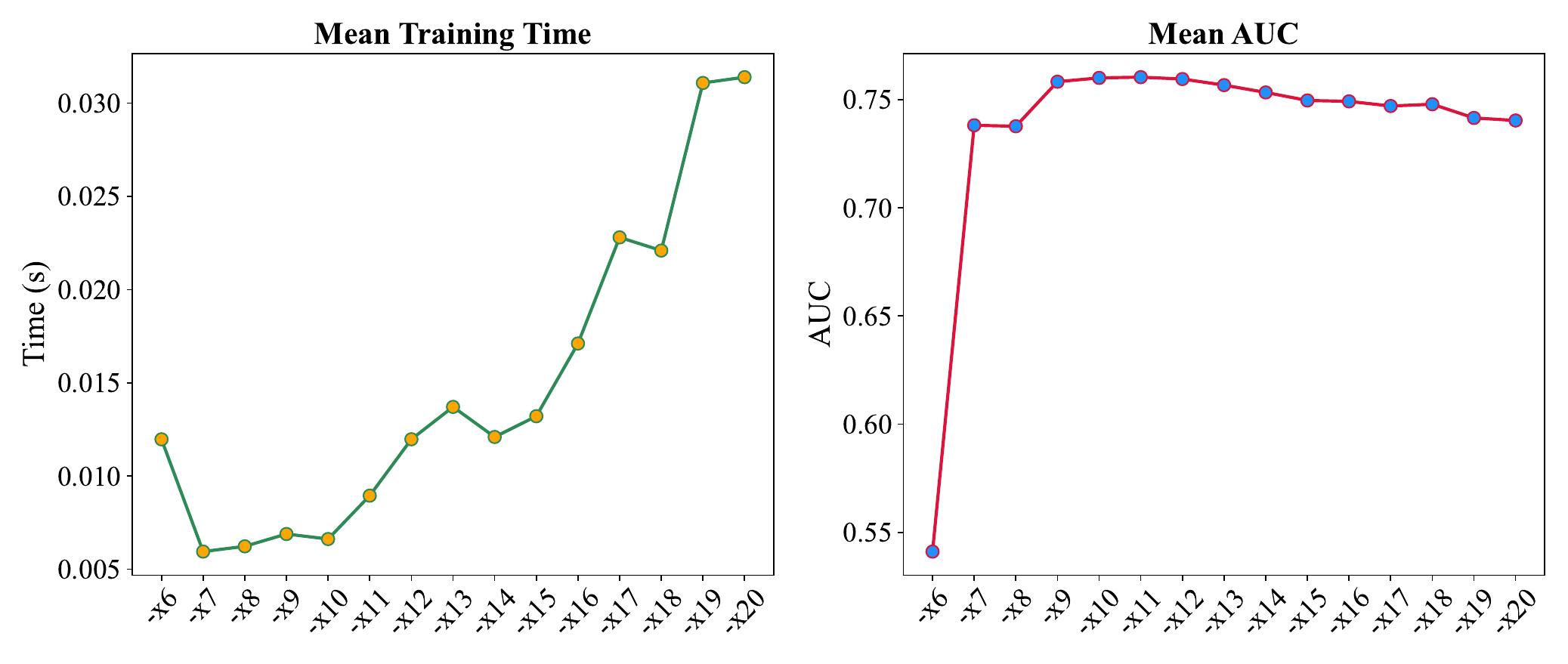}
		\label{subfig: bitcon_data_x6tox20_p3}} 
	\setlength{\subfigcapskip}{-0.5em}
	\subfigure[\texttt{Target\_$5$period}]{\includegraphics[scale=0.3]{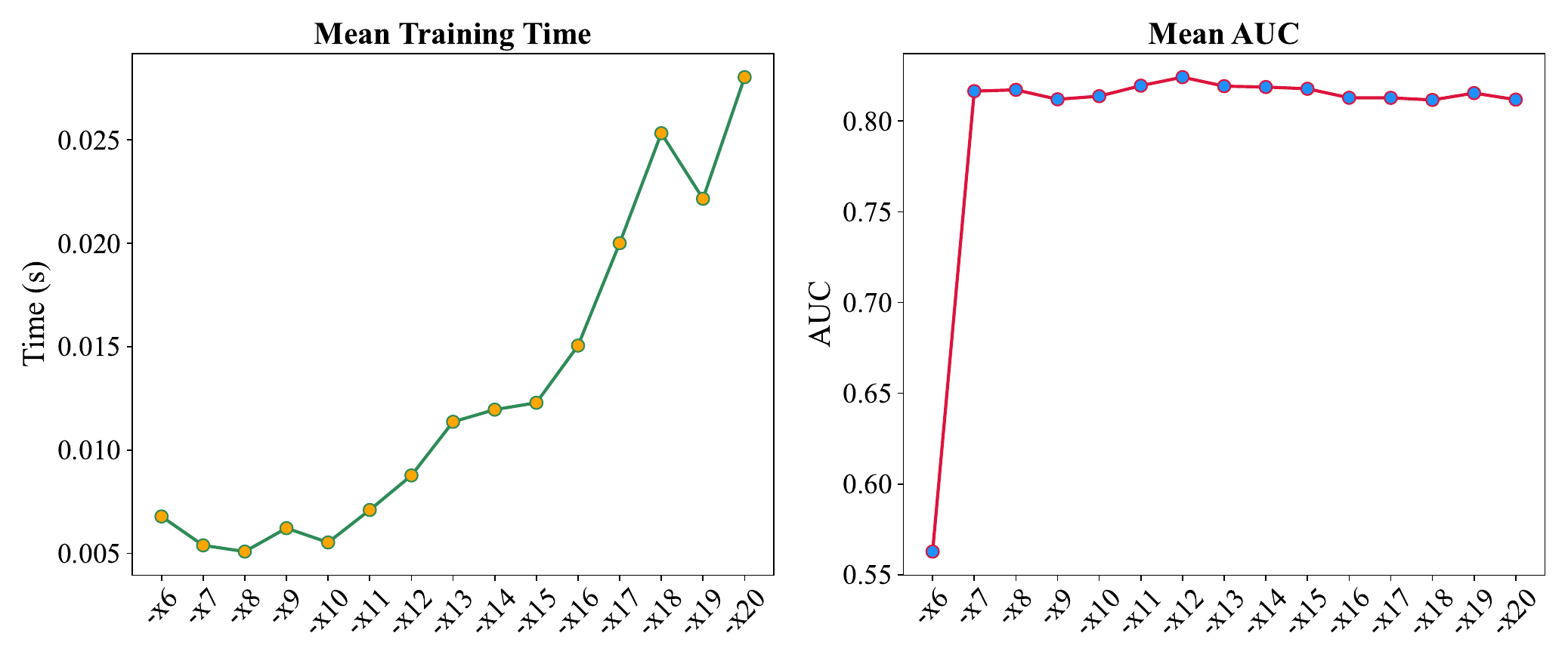}
		\label{subfig: bitcon_data_x6tox20_p5}} 
        	\caption{Importance of the lagged time-related target variables}
	\label{fig: bitcon_data}
\end{figure}

Figure \ref{fig: bitcon_data} shows that for both forecasting horizons, the mean AUC exhibits a rapid initial improvement followed by a very slight decline, confirming the value of historical price information in classifying price directions. Notably, this improvement in accuracy is achieved without increasing computational cost, demonstrating the computational efficiency of IPL.

\vspace{-0.5cm}

\begin{table}[htbp]
	\centering
	\caption{Classification performance comparison for \texttt{Target\_$3$period}}
	\label{tab:performance_comparison1}
    	\scalebox{0.77}{
        \begin{tabular}{p{3.6cm} p{3.8cm} cccc}
		\toprule
		\multicolumn{1}{l}{ \multirow{2}*{Features} }   & \multicolumn{1}{l}{ \multirow{2}*{Methods}}   & \multicolumn{3}{c}{\textit{Class 1 Performance}} & \multicolumn{1}{c}{ \multirow{2}*{\textit{Accuracy}} }  \\
		\cmidrule(lr){3-5}
		\multicolumn{1}{c}{}& \multicolumn{1}{c}{}& \textit{Precision} & \textit{Recall} & \textit{F1-score} & \multicolumn{1}{c}{} \\
		\midrule
		\multirow{4}{*}{Top 3} 
        & LIME & 0.73 & 0.51 & 0.60 &0.64   \\
        & SHAP & \textcolor{blue}{0.76} & 0.56 & 0.65 & 0.69 \\
        & ARIMAX (2,0,2) & 0.58 & 0.53 & 0.55 & 0.53 \\
		& \textbf{IPL} & 0.75 & \textcolor{blue}{0.74} & \textcolor{blue}{0.74} & \textcolor{blue}{0.73} \\
		\addlinespace
        \addlinespace
		\multirow{4}{*}{Top 1} 
          & LIME & 0.72  & 0.71 &  0.71&  0.70  \\
        & SHAP & 0.74 & 0.55 & 0.63 & 0.68 \\
        & ARIMAX (2,0,2) & 0.55 & 0.51 & 0.53 & 0.50 \\
		& \textbf{IPL} & \textcolor{blue}{0.75} & \textcolor{blue}{0.74} & \textcolor{blue}{0.74} & \textcolor{blue}{0.73} \\
		\bottomrule
	\end{tabular}}
        \vspace{3pt} 
		\small
		\begin{tabular}{@{}l@{}}
			\multicolumn{1}{@{}p{0.75\linewidth}@{}}{
				\scriptsize
				\textbf{Note}: 
				\textit{Precision} = $\frac{TP}{TP + FP}$, \textit{Recall} = $\frac{TP}{TP + FN}$,  \textit{F1-score} = $\frac{2 \times \textit{Precision} \times \textit{Recall}}{\textit{Precision} + \textit{Recall}}$, and \textit{Accuracy} = $\frac{TP + TN}{TP + FN + TN + FP}$, where $TP$, $TN$, $FP$, and $FN$ denote true positive, true negative, false positive, and false negative, respectively. The best results  are highlighted in color.

			}
		\end{tabular}
\end{table}
\vspace{-0.5cm}

\subsection{Advantages of IPL in identifying important features}

Table \ref{tab:performance_comparison1} compares the performance of widely used interpretable decision tree models built on features identified by different interpretability methods, evaluated under common metrics on cryptocurrency price data \citep{zhang2024generalized}. The results show that models using IPL-identified features consistently achieve higher accuracy than those using features identified by LIME, SHAP, or ARIMAX, regardless of whether the top three features or only the single most important feature is considered. These results indicate that IPL provides more  reliable feature-level interpretability.

\section{Field experiment: Antenna health management}

Antenna maintenance personnel continuously monitor data from both the antenna system and environmental sensors to assess operational status and perform anomaly detection for predictive maintenance. Consequently, it is essential to develop a machine learning–based early warning mechanism that achieves high prediction accuracy while maintaining feature-level interpretability.
The dataset used in this study consists of 5,437 samples collected from physical antenna equipment at the 39th Research Institute of China Electronics Technology Group Corporation. The data are chronologically divided into training (3,437 samples), validation (1,000 samples), and test sets (1,000 samples), with the most recent samples assigned to the test set.
A detailed description of the dataset variables is provided in Table~\ref{tab:dataset_description}.

\begin{table}[htbp]
            \caption{Description of the field-collected antenna data}
            \vspace{-0.5cm}
    	\begin{center}
	\scalebox{0.77}{
\begin{tabular}{ccc}
\toprule
\textbf{Variable} & \textbf{Description} &\textbf{ Range / Note} \\
\midrule
$x_1$ & Position angle value & $0^\circ \sim 360^\circ$, obtained by combining $x_4$ and $x_5$ \\
$x_2$ & Speed & $0 \sim 2000$ r/min, collected by tachometers \\
$x_3$ & Current ratio & $0 \sim 100\%$, ratio of collected current to rated current \\
$x_4$ & Coarse code value & $0 \sim 65535$, 16-bit binary, resolver feedback \\
$x_5$ & Fine code value & $0 \sim 65535$, 16-bit binary, resolver feedback \\
$x_6$ & Label & $-1$ = normal, $1$ = abnormal \\
\bottomrule
\end{tabular}}
    \vspace{3pt} 
		\small
		\begin{tabular}{@{}l@{}}
			\multicolumn{1}{@{}p{0.76\linewidth}@{}}{
				\scriptsize
				\textbf{Note}: 
				Given the circular nature of the angular variable $x_1$, it was transformed into trigonometric representations: $x_{1\text{sin}} = \sin(2\pi x_1)$ and $x_{1\text{cos}} = \cos(2\pi x_1)$.
			}
		\end{tabular}
        	\end{center}
\label{tab:dataset_description}
\end{table}
	\vspace{-0.3cm}



\begin{figure}[H]
	\centering
	\vspace{0.1in}
	\includegraphics[scale=0.35]{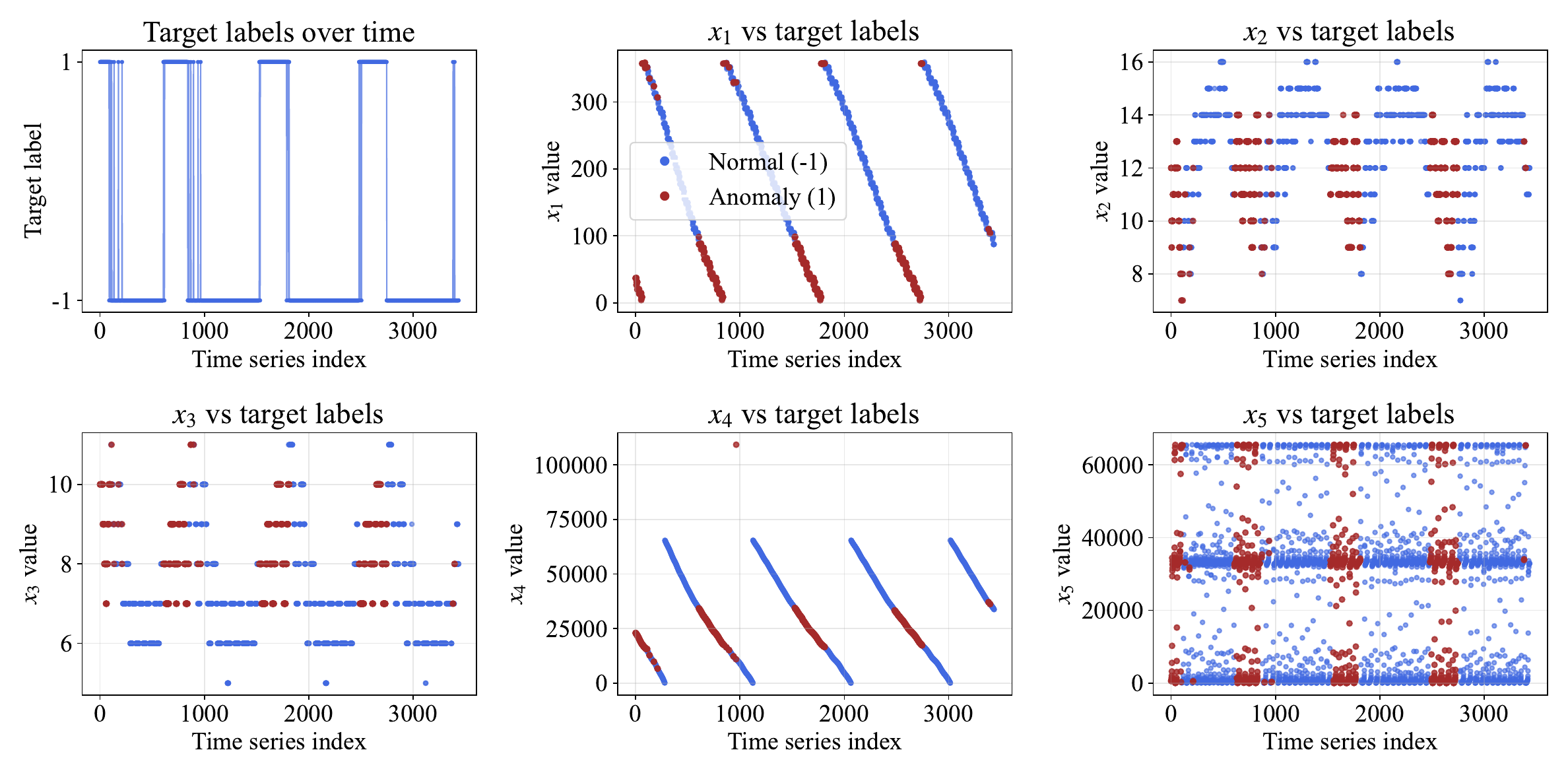}
    	\caption{Distribution of target labels across different feature values}
	\label{Fig:data1_describe}  
	\vspace{-0.3cm}
\end{figure}

Figure \ref{Fig:data1_describe} provides an initial visualization of the dataset, highlighting two key observations. First, the anomaly labels exhibit distinct periodic patterns and tend to occur consecutively over specific time intervals. Second, only the position angle and coarse code values demonstrate clear separability between normal and abnormal states. However, neither the position angle nor the coarse code alone is sufficient for accurate early warning, and the lack of discriminability in other features highlights the necessity of analyzing higher-order feature interactions for better early warning.

\vspace{-0.4cm}

We conducted two sets of experiments under the hinge loss. First, we verify the effectiveness of IPL in providing feature interpretability by examining the relationship between feature sparsity and prediction accuracy. Second, we compare different interpretability methods in designing an early warning mechanism to demonstrate the advantages of IPL.
For fair comparison, all interpretability methods are first tuned to achieve similarly high accuracy ($>$ 0.95).
 Given the similarity in setup and visualization to prior experiments, the corresponding feature importance rankings are included in the Appendix for reference.

\subsection{Feature sparsity-accuracy analysis for evaluating interpretability}

\begin{figure}[H]
	\centering
	\vspace{0.1in}
	\setlength{\subfigcapskip}{-0.5em}
	\subfigure[Input using only the original variables]{\includegraphics[scale=0.26]{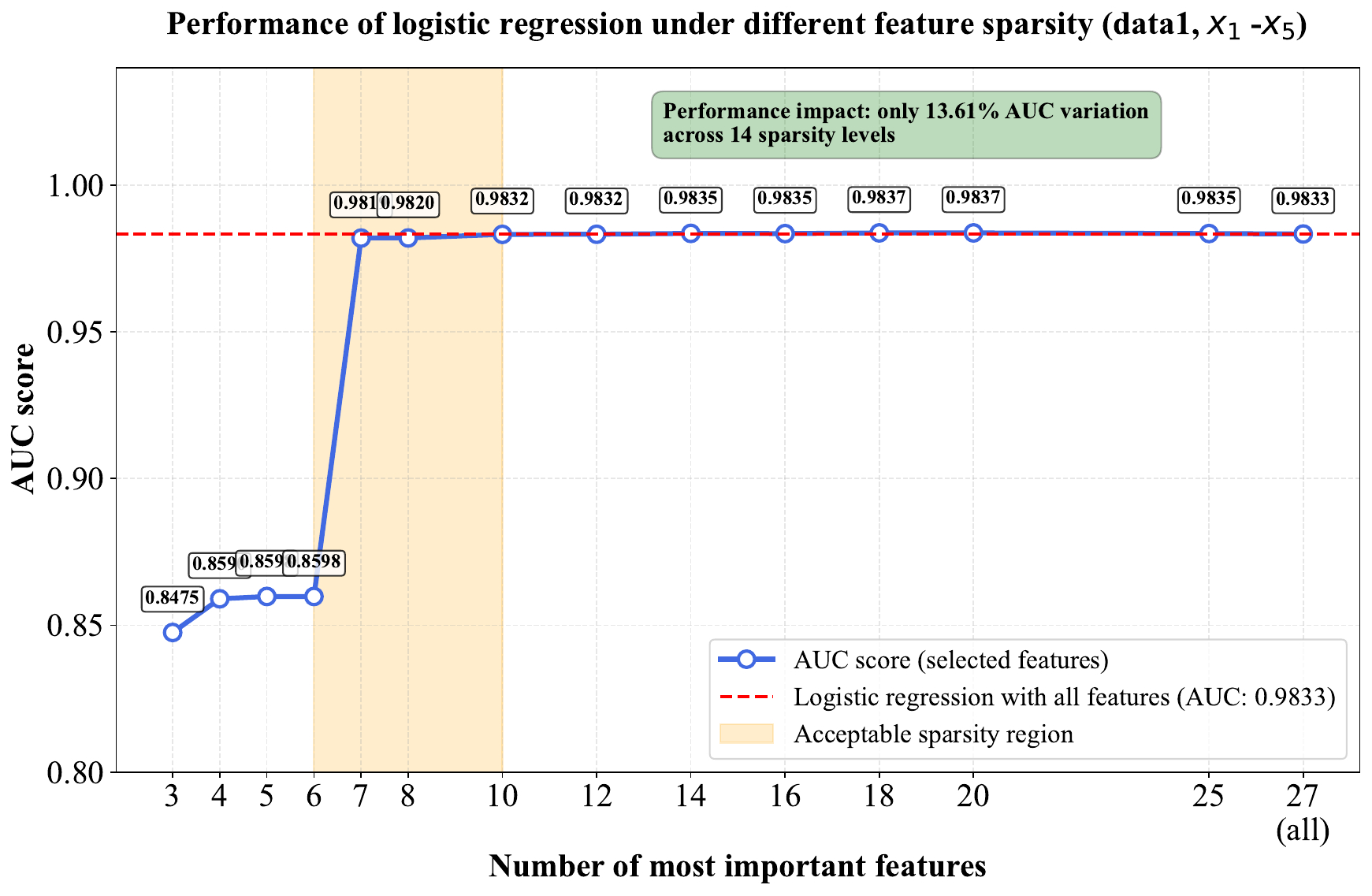}
		\label{subfig: sparsity_auc_plot_data1_d5}} 
	\setlength{\subfigcapskip}{-0.5em}
	\subfigure[Input with original variables and three lagged target labels]{\includegraphics[scale=0.26]{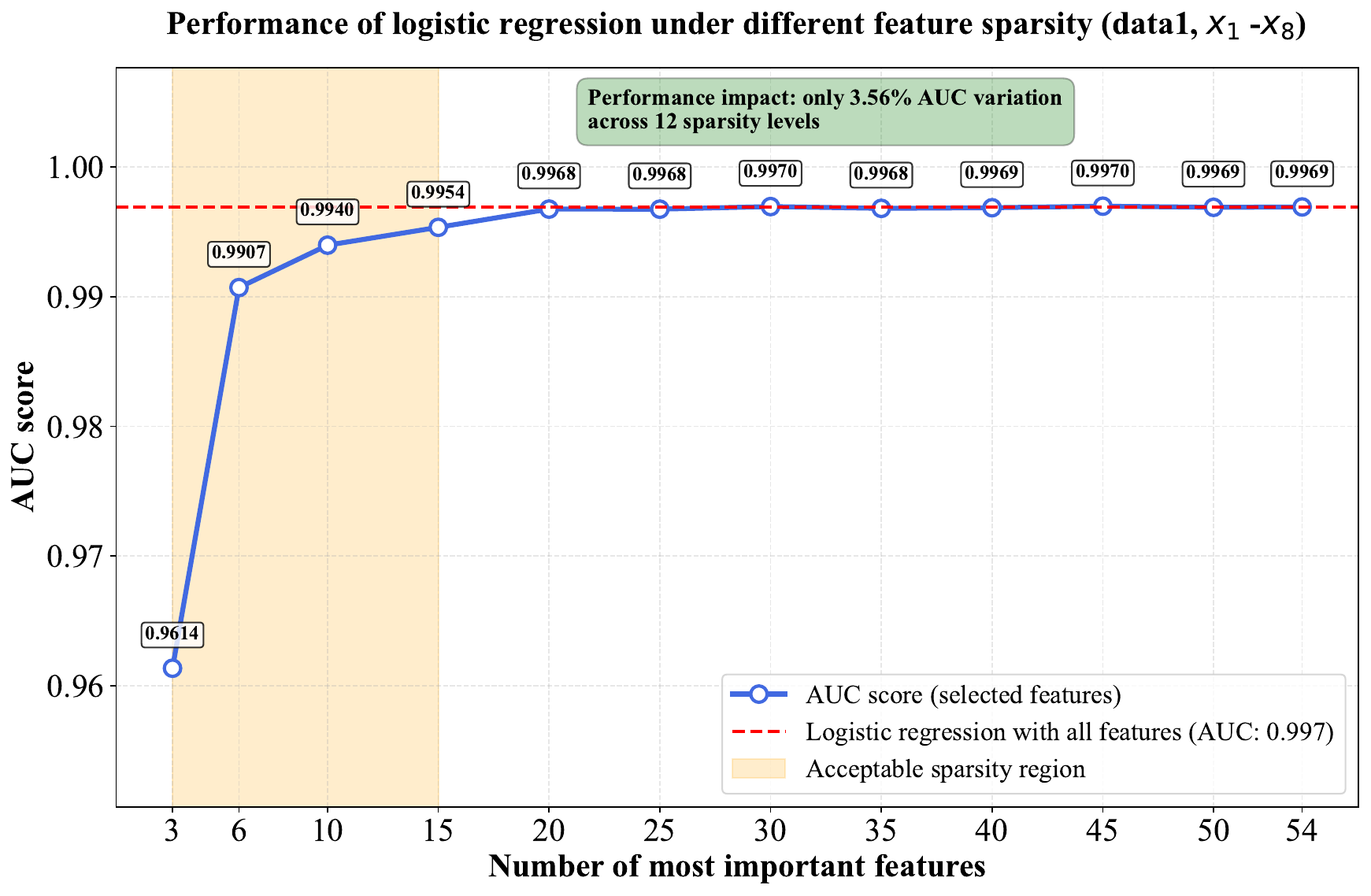}
		\label{subfig: sparsity_auc_plot_data1_d10}} 
        	\caption{Relationship between IPL based feature sparsity and prediction performance}
	\label{fig: sparsity_auc_plot_data1}
\end{figure}

This experiment investigates the relationship between the number of top-ranked features selected by IPL and the resulting classification accuracy, to evaluate IPL's effectiveness in providing feature interpretability. To ensure a fair comparison, a logistic regression classifier is used for prediction across all sparsity levels. The results are shown in Figure \ref{fig: sparsity_auc_plot_data1}, which includes performance under two input conditions: using only the originally collected features, and adding time-related target variables as part of the input.

The light yellow region indicates the range of sparsity within which a reduced feature set maintains relatively high accuracy. Based on this region, we conclude that data collectors can achieve classification accuracy comparable to the full feature set by focusing on only 7 to 10 top-ranked features.
Furthermore, the variation in AUC across different sparsity levels is minimal, with a maximum fluctuation of only 13.61\%. These results confirm that IPL effectively provides feature interpretability by successfully identifying and ranking the most influential features.

\vspace{-0.5cm}

\subsection{Feature-interpretability-based early warning mechanism design}

In this section, we design early warning mechanisms based on the feature importance derived from different interpretability methods. Given the binary classification nature of our problem, we adopt a binary tree structure to construct each early warning mechanism.

\begin{figure}[!t]
	\centering
	\vspace{0.1in}
	\includegraphics[scale=0.55]{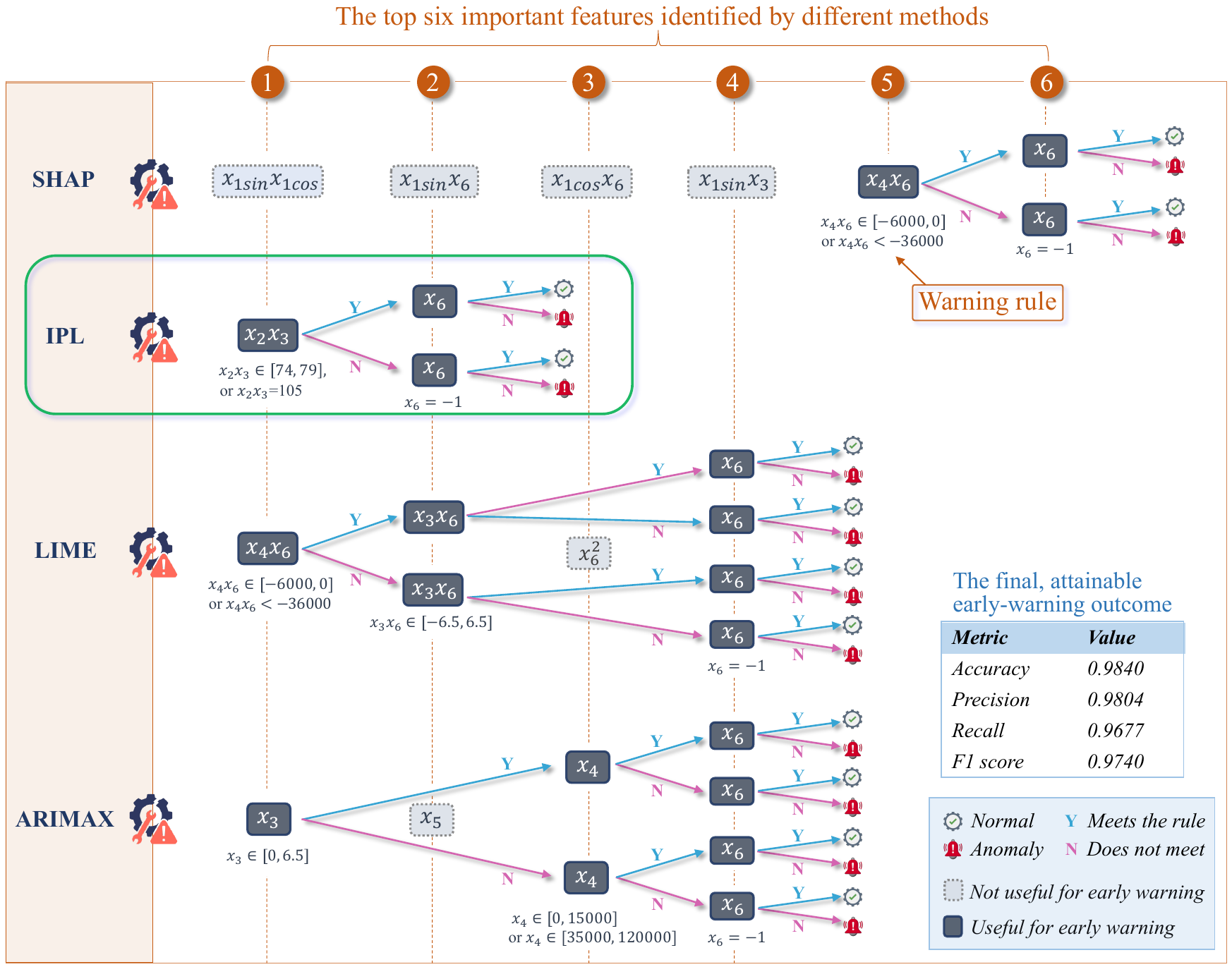}
    	\caption{Early-warning mechanisms designed using feature importance rankings derived from different methods}
	\label{Fig:early warning compare}  
	\vspace{-0.3cm}
\end{figure}

Figure \ref{Fig:early warning compare} ranks the features of each method in descending order of importance and presents the corresponding early warning mechanisms designed to achieve satisfactory performance (with each metric exceeding 0.96). In the figure, light-gray rectangles represent features that do not contribute effectively to early warning, meaning they lack a clear discriminant boundary between normal and abnormal states, while dark-gray rectangles denote the decision nodes that constitute the warning rules. Specific warning rules are displayed below each dark-gray node.
Two main observations can be drawn. (1) The IPL-based warning mechanism is notably simpler and more efficient, requiring only two features, which are precisely the top two ranked in importance. This demonstrates IPL’s advantage in feature interpretability: it accurately identifies the features that contribute most to predictions. (2) The most important feature identified by IPL, $x_2x_3$, corresponds to the product of antenna rotational speed and current ratio. This interaction term effectively captures abnormal patterns because it approximates mechanical power, and anomalous states often manifest as power anomalies. Additionally, among the important interaction terms identified by LIME and SHAP, $x_6$ is consistently present, reflecting the influence of past states on the current state.

\begin{figure}[!t]
	\centering
	\vspace{0.1in}
	\includegraphics[scale=0.57]{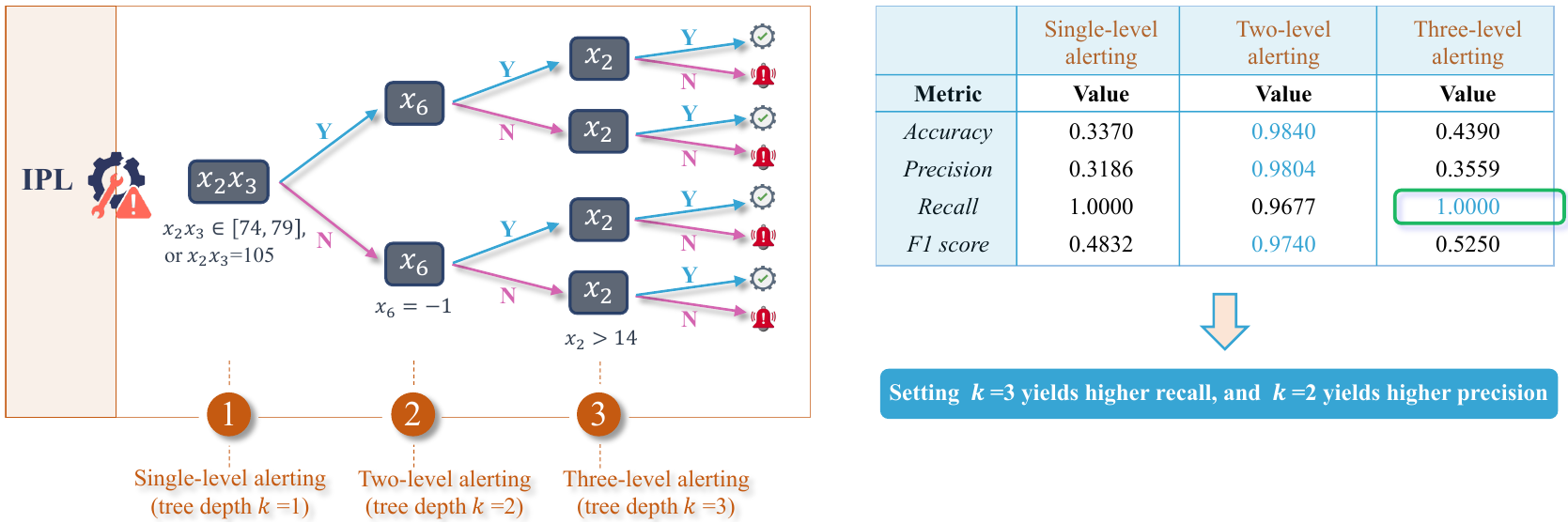}
    	\caption{Early-warning results based on IPL with varying tree depths ($k$)}
	\label{Fig:early warning IPL}  
	\vspace{-0.3cm}
\end{figure}

In antenna health management, beyond conventional metrics such as accuracy or precision, greater emphasis is placed on the early detection of abnormal states—particularly the ability to trigger alerts during prolonged normal operation that signal potential future anomalies. Therefore, we provide a detailed illustration of the IPL-based early warning mechanism in Figure~\ref{Fig:early warning IPL}. Administrators may adopt a mechanism with tree depth $k=2$ to achieve higher precision, or choose a deeper tree ($k=3$) to attain perfect recall (1.0). Although the latter slightly reduces overall precision, this trade-off is often acceptable in antenna maintenance, where missed anomalies incur high operational costs.

While the $k=2$ mechanism achieves a recall of 0.9677, this performance is partly attributable to the temporal continuity inherent in antenna operational states. Upon closer inspection of the test samples, we observe that all instances corresponding to transitions from normal to abnormal states are missed by the $k=2$ mechanism but are successfully captured by the $k=3$ mechanism. Several representative samples are provided in Table~\ref{Tab:early warning}, which further demonstrate the mechanism’s ability to issue warnings across consecutive future time steps, highlighting its practical value for long-term state forecasting.

\begin{table}[!htbp]
\centering
\caption{Justification for setting $k = 3$: triggering alarms for anomalies during sustained normal operation}\label{Tab:early warning}
\scalebox{0.76}{
\begin{tabular}{ccccccccccc}
\toprule
\multirow{3}*{\makecell[c]{Sample\\index}}
 & \multicolumn{3}{c}{\textbf{Early warning features}} & \multicolumn{3}{c}{\textbf{Labels at  the current time} $t_1$} & \multicolumn{3}{c}{\textbf{Labels at the next time} $t_2$} &  \multirow{3}*{$T_n$} \\
\cmidrule(lr){2-4} \cmidrule(lr){5-7} \cmidrule(lr){8-10}
 & \multirow{2}*{$x_2x_3$}  &\multirow{2}*{$x_6$}  & \multirow{2}*{$x_2$} & \textit{Predicted} & \textit{Predicted} & \textit{Ground} & \textit{Predicted} & \textit{Predicted} & \textit{Ground} & \\
 &  &  &  & $(k=2)$ & \textcolor{blue}{$(k=3)$} & \textit{truth} & $(k=2)$ & \textcolor{blue}{$(k=3)$} & \textit{truth} & \\
\midrule
\textit{5034, 5048} & 91.0 & -1 & 13.0 & -1 & \textcolor{blue}{1} & \textcolor{blue}{1}  & -1 & \textcolor{blue}{1}  & \textcolor{blue}{1}  & 2 \\
\textit{5065, 5082} & 96.0 & -1 & 12.0 & -1 & \textcolor{blue}{1}  & \textcolor{blue}{1}  & -1 & \textcolor{blue}{1}  & \textcolor{blue}{1}   & 10\\
\bottomrule
\end{tabular}}
 \vspace{3pt} 
		\footnotesize
		\begin{tabular}{@{}l@{}}
			\multicolumn{1}{@{}p{0.96\linewidth}@{}}{
				\scriptsize
				\textbf{Note}: $T_n$ represents the number of consecutive future time steps that can be accurately predicted. $k$ denotes the tree depth of the early‑warning mechanism.}
		\end{tabular}
\end{table}

\section{Conclusion}

Time-series data play a critical role in practical applications by enabling the prediction of future system states based on historical observations. One of the most important application domains is early warning, such as predictive maintenance in engineering systems and health monitoring in medical settings, where timely warnings can substantially reduce resource consumption and operational costs. Therefore, it is essential to develop high-precision early warning mechanisms that are interpretable by human operators and support direct intervention at the level of original observed variables. This motivates the need for time-series forecasting methods that can simultaneously achieve high prediction accuracy and feature-level interpretability.

Existing intrinsically interpretable methods (e.g., ARIMAX) can provide feature-level explanations, but they generally suffer from limited prediction accuracy. Widely used post-hoc explanation techniques, such as LIME and SHAP, are also inadequate for interpreting time-series data, as they often neglect the inherent temporal dependencies among observations. Although numerous deep learning–based methods have been proposed for interpretable time-series forecasting, they primarily focus on temporal importance rather than providing feature-level interpretability of the original input variables, making it difficult to translate their explanations into actionable guidance for early warning. Moreover, such methods are often computationally intensive and difficult for non-experts to use.
In summary, there remains a lack of time-series–specific forecasting methods that can simultaneously achieve high prediction accuracy and feature-level interpretability to support early-warning applications. To address this gap, we propose the IPL method, which tightly couples features with the model through a polynomial structure, explicitly incorporating features from different time steps as well as their interaction terms into the model. IPL preserves the inherent temporal structure of time-series data and enables a flexible balance between accuracy and interpretability by adjusting the polynomial order, allowing practitioners to select a model that maintains high prediction accuracy while ensuring feature-level interpretability.

We validate the advantages of IPL in terms of the accuracy–interpretability trade-off by comparing it with widely used interpretability methods, including LIME, SHAP, and ARIMAX, on both simulated data and two real-world datasets. On the field-collected antenna dataset, we further compare early warning mechanisms constructed based on the feature importance identified by different methods. The results show that the interpretability provided by IPL enables the construction of a more streamlined early warning mechanism while achieving performance comparable to that of other methods. Moreover, based on the derived feature importance, the depth of the warning mechanism can be flexibly adjusted to meet different precision requirements.

Our approach is applicable to a wide range of application domains. In finance, for example, IPL can help identify the key factors driving commodity price fluctuations, thereby enabling proactive risk management. In healthcare, it can provide medical staff with precise and interpretable health alerts, supporting targeted interventions while mitigating alert fatigue. In engineering equipment management, IPL can assist managers in anticipating maintenance needs and scheduling inspections in advance, thereby addressing the growing demand for predictive maintenance.

\bibliographystyle{elsarticle-num-names} 
\bibliography{ref}

\end{document}